\documentclass{article} 
\usepackage{iclr2027_conference,times}
\iclrfinalcopy 

\usepackage{amsmath,amsfonts,bm}

\def\eqref#1{equation~\ref{#1}}

\def\1{\bm{1}}

\DeclareMathAlphabet{\mathsfit}{\encodingdefault}{\sfdefault}{m}{sl}
\SetMathAlphabet{\mathsfit}{bold}{\encodingdefault}{\sfdefault}{bx}{n}

\usepackage{hyperref}
\usepackage{url}
\usepackage{graphicx}
\usepackage{booktabs}
\usepackage{multirow}
\usepackage{xcolor}
\usepackage{subcaption}
\usepackage{amsmath}
\newcommand{\reslogo}[1]{\raisebox{-0.22em}{\includegraphics[height=1.15em]{#1}}}
\newcommand{\reslink}[2]{\href{#1}{\textcolor{blue!45!black}{#2}}}

\graphicspath{{figures/}}

\newcommand{\ours}{Nautil}
\newcommand{\sft}{Nautil-SFT}
\newcommand{\rlvr}{Nautil-RLVR}
\newcommand{\closed}{\textsc{case closed}}
\newcommand{\notclosed}{\textsc{case not closed}}

\title{Not Until the Evidence Says So: Teaching LLM Investigators When to Close a Case}

\author{Tingzhu Bi \\
Peking University \\
\texttt{bitingzhu@stu.pku.edu.cn} \And
Ping Wang \\
Peking University \\
\texttt{pwang@pku.edu.cn} \And
Meng Ma \\
Peking University \\
\texttt{mameng@pku.edu.cn}}

\begin{document}

\maketitle

\vspace{-1.6em}
\begin{center}\small\hypersetup{pdfborder={0 0 0}}
\reslogo{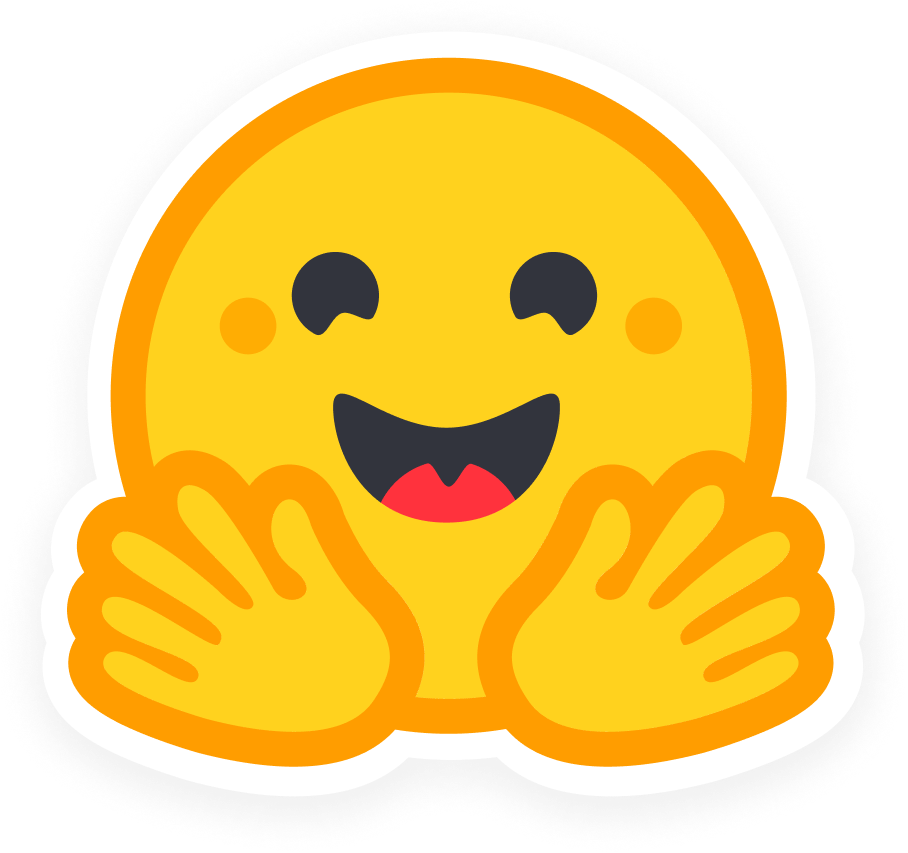}~\textbf{Dataset:} \reslink{https://huggingface.co/datasets/etigerstudio/Nautil}{Nautil}
\quad
\reslogo{hf-logo.png}~\textbf{Models:} \reslink{https://huggingface.co/etigerstudio/Nautil-SFT}{Nautil-SFT}, \reslink{https://huggingface.co/etigerstudio/Nautil-RLVR}{Nautil-RLVR}
\quad
\reslogo{hf-logo.png}~\textbf{Demo:} \reslink{https://huggingface.co/spaces/etigerstudio/Nautil-Demo}{Nautil-Demo}
\quad
\reslogo{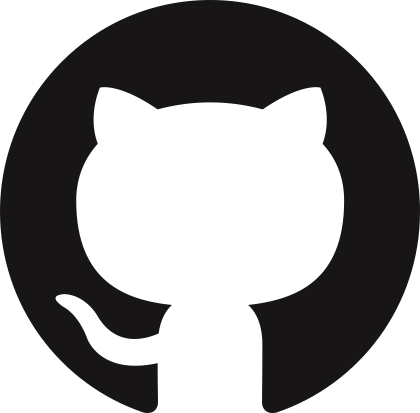}~\textbf{Code:} \reslink{https://github.com/etigerstudio/Nautil}{Nautil}
\end{center}
\vspace{0.4em}

\begin{abstract}
Accident, defect and outage investigations end with a decision that ordinary question answering never faces: whether the evidence gathered so far is enough to close the case.
We study this decision for LLM investigators, which request evidence from a case file, revise their hypotheses, and either close the case with a conclusion grounded in what they read or leave it open and name what is missing.
This judgment does not come with capability: an untrained 9B model overstates its evidence in 97\% of its answers, and a frontier model that identifies the right cause in 84\% of cases still overstates in 91\% and closes 17 of the 41 cases whose official finding is ``cause undetermined''.
Measuring it is also non-trivial: the source of a case largely predicts its label, and a rule that reads only the source reaches 83.0 balanced accuracy on our test cases.
We therefore evaluate closure with three tests: \emph{closure accuracy}, reported against this rule and within each source; \emph{evidence dependence}, which removes the grounds of a conclusion and checks whether the model stops closing; and \emph{conclusion and gap quality}, a judged checklist of what the model asserts and what it says is missing.
We build \ours{}, 731 audited cases from aviation, rail, maritime, chemical-safety and vehicle-defect reports and production server incidents, with teacher trajectories, an out-of-distribution test set and counterfactual evidence versions.
Fine-tuning a 9B model on these trajectories makes its closures follow the evidence: removing the grounds lowers its closure rate by 26 points relative to a matched control (base model: 6; frontier model: 7), overstatement falls from 97\% to 35\%, and conclusions that are both correct and not overstated rise from 3\% to 43\% (frontier model: 9\%).
Reinforcement learning that rewards only the closure decision then raises balanced accuracy from 69.2 to 83.3, on par with the teacher, and within-source accuracy from 60.4 to 74.1, at some cost in evidence dependence.
We release the models, the dataset and the evaluation suite.
\end{abstract}

\section{Introduction}
\label{sec:intro}

\begin{figure*}[t]
\centering
\includegraphics[width=\textwidth]{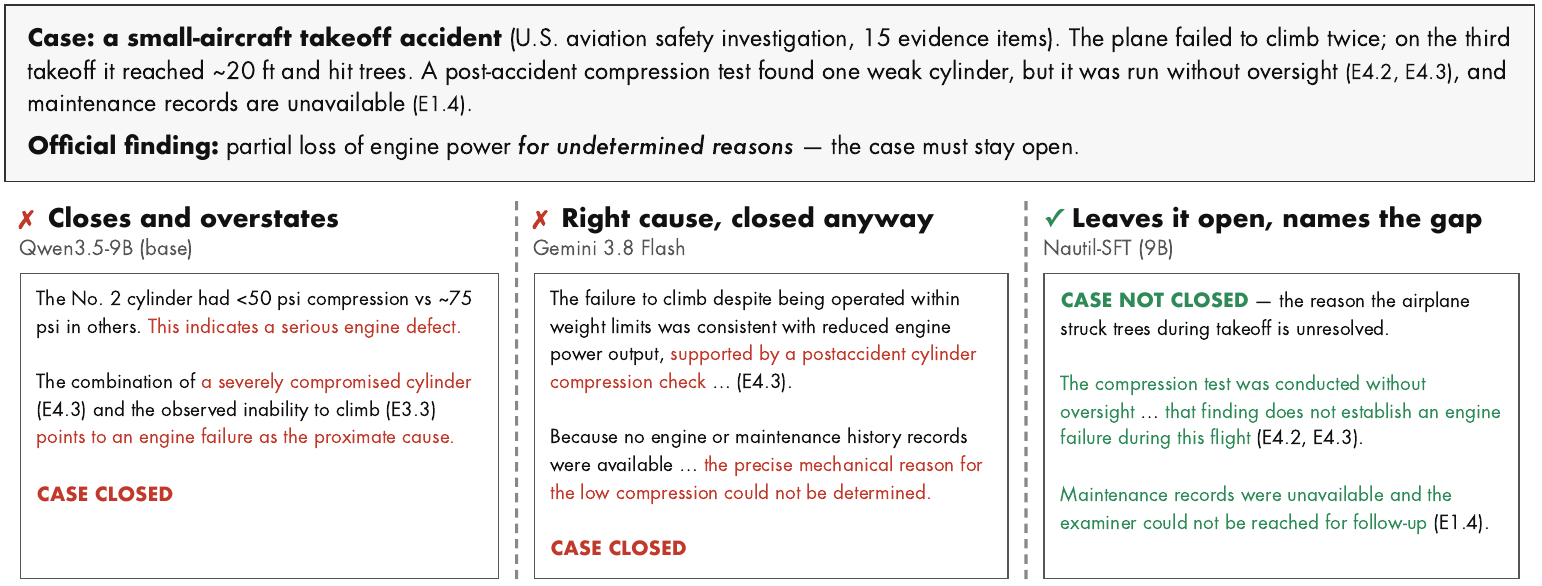}
\caption{Three investigators on a test case whose official finding is a partial loss of engine power \emph{for undetermined reasons}. Left: the untrained base model turns an unverified reading into an established engine failure and closes the case. Middle: a frontier model names the power loss that the report establishes and even writes that its cause could not be determined, yet still closes the case. Right: \sft{} keeps the case open and names the missing evidence. Excerpts are the models' own words, shortened (\ldots); one sample per model.}
\label{fig:caseA}
\vspace{-4pt}
\end{figure*}

Investigating an aircraft accident, a vehicle defect or a server outage means collecting evidence piece by piece and then deciding whether it is enough to state a cause.
Official investigations often decide that it is not: many reports end with ``cause undetermined''.
An LLM investigator therefore has to make two decisions, not one: what the cause is, and whether the evidence at hand establishes it.

Current models are much better at the first decision than at the second (Figure~\ref{fig:caseA}).
On our test cases, a frontier model (Gemini 3.8 Flash) identifies the reference cause in 84\% of cases, yet 91\% of its answers present as established what the evidence only suggests, and it closes 17 of the 41 cases that the official reports leave undetermined.
An untrained 9B model overstates in 97\% of its answers.
Knowing when to close does not come with the ability to find causes; it has to be taught and measured separately.

Abstention research asks whether to answer a single question whose information is given up front \citep{kirichenko2026abstentionbench,zhai2026abstainr1}, and diagnosis agents are scored on finding the cause \citep{xu2025openrca,pham2025rcaeval}, not on whether they should have concluded.
Our cases hold tens to hundreds of evidence items that must be requested over several turns, and ``the cause cannot be determined from this record'' is a correct, scored outcome.

Measuring this decision is harder than it looks.
In real investigation data the source of a case largely predicts its label: most NTSB and NHTSA cases in our data end undetermined, most board and server cases end with a cause.
A rule that ignores the evidence and answers by source reaches 83.0 balanced accuracy on our non-host test cases.
Any closure accuracy on such data must therefore be read against this rule, within each source, and together with tests that change the evidence while keeping the source fixed.

We make four contributions.
\begin{itemize}
\item \textbf{Task and evaluation.} We define investigative closure and evaluate it with three complementary tests: closure accuracy (reported against the source-only rule and within each source), evidence dependence (counterfactual removal of a conclusion's grounds against a matched control), and conclusion and gap quality (a judged checklist).
\item \textbf{Data.} \ours{}\footnote{Read as \emph{not until}.} contains 731 audited cases from four sources, including real production server incidents, with teacher investigation trajectories, a 141-case out-of-distribution test set and counterfactual evidence versions.
\item \textbf{Models.} \sft{} and \rlvr{}, two 9B investigators. Supervised fine-tuning on audited trajectories makes closure follow the evidence and cuts overstatement; reinforcement learning with a reward on the closure decision alone raises closure accuracy to the teacher's level.
\item \textbf{Findings.} Training teaches restraint and grounding rather than cause-finding; rewarding judged conclusion quality during RL collapses the policy into never closing; and an outcome-only reward trades evidence dependence for accuracy.
\end{itemize}

\section{Related Work}
\label{sec:related}

\paragraph{Abstention and unanswerable questions.}
Models answer questions they cannot answer \citep{rajpurkar2018know,yin2023large,kadavath2022language}, even when the context is insufficient \citep{joren2025sufficient,kirichenko2026abstentionbench}, partly because binary grading rewards guessing \citep{kalai2025why}.
Abstention has been trained with SFT \citep{zhang2024rtuning,yang2024alignment} and RL \citep{wei2026truthrl,damani2026beyond,zhai2026abstainr1}; RL fine-tuning can also erode it \mbox{\citep{song2025hallucination}}, and abstention rewards can collapse into refusal \citep{che2026abstention}.
Recent work asks models to state what is missing \citep{zhai2026abstainr1,li2026questbench} or to stop acting when they should \citep{liu2026agentabstain,luo2026agentic,liu2025cart}.
These settings give the information up front and grade whether to answer; in ours, the decision follows multi-turn evidence gathering, and ``cannot determine'' is set by real investigation reports.

\paragraph{LLMs for incident investigation and diagnosis.}
Root-cause analysis benchmarks and agents for cloud incidents \citep{xu2025openrca,pham2025rcaeval,chen2025aiopslab,wang2024rcagent}, auditable multi-agent diagnosis such as JustDiag \citep{bi2026justdiag}, evidence-first investigators \citep{marozzo2026llm,chen2026holmes}, and sequential medical diagnosis \citep{nori2025sequential,li2024mediq} all score whether the cause is found. Premature closure has been measured in frontier models \citep{handler2026quantifying}. We add closure itself as a scored, trainable decision across domains.

\paragraph{RLVR and agentic RL.}
Group-relative policy optimisation \citep{shao2024deepseekmath,guo2025deepseekr1} and tool-using RL agents \citep{jin2025searchr1,song2025r1searcher,wang2025ragen} assume a solvable task with a checkable answer. We make ``not closed'' one of the verifiable outcomes and show that judge-scored rubric rewards \citep{gunjal2026rubrics}, which can be gamed \citep{skalse2022defining,gao2023scaling}, collapse closure in this setting.

\paragraph{Shortcuts and faithful evaluation.}
Annotation artefacts and shortcuts let models score without using the input \citep{gururangan2018annotation,geirhos2020shortcut}; contrast sets and counterfactual edits test whether predictions follow it \citep{gardner2020evaluating,kaushik2020learning}, though constructed insufficiency has its own artefacts \citep{qiu2026evidence}. LLM judges are useful but biased, including towards their own outputs \citep{zheng2023judging,panickssery2024llm}. These motivate our source-only baseline, matched control and cross-judge checks.

\section{Investigative Closure}
\label{sec:task}

\paragraph{Protocol.}
An investigator receives a case: a fixed question (what caused the incident, citing evidence for each key claim), a brief, an opening record $E_1$ shown in full (median 4 items), and an index that lists only the identifier and a neutral title of every other evidence item (over all 731 cases: median 115 items per case, up to 682).
Titles are written so that they do not reveal conclusions.
In each turn the investigator writes a working note (current hypotheses, what the new evidence supports or weakens, a hypothesis ledger, and the next discriminating check) and may call \texttt{request\_evidence} for 1--12 unread items; the items arrive in the next message.
An episode has at most 15 turns and ends with an answer that starts with \closed{}, followed by a causal conclusion that cites the evidence, or \notclosed{}, followed by what is established, what is missing, and a provisional assessment.
The label of a report case is whether the official report determined the cause at the level its investigation pursued (the closure object in the reference); a report that establishes an event, such as a loss of engine power, but not its cause leaves the case open.

\paragraph{Three tests.}
A single accuracy can be inflated either by the source of a case or by a conservative policy, so we evaluate closure with three tests, each closing one loophole: \emph{closure accuracy} guards against recognising the source instead of reading the evidence; \emph{evidence dependence} guards against closing out of habit, or stopping merely because evidence is scarce; and \emph{conclusion and gap quality} guards against a right decision with a wrong or overstated conclusion.

\emph{Closure accuracy.}
We report the rate of closing among cases that should close, the rate of not closing among cases that should not, and their mean (balanced accuracy), so that never closing is not rewarded. An answer without a closure marker counts as wrong on both sides.
Every result is shown next to the \emph{source-only rule}, which gives each source its majority training label.
\emph{Within-source balanced accuracy} computes balanced accuracy separately inside each report source and averages them; the source-only rule scores exactly 50 on it, so only reading the evidence can raise it.

\emph{Evidence dependence.}
For cases labelled closed, we build a version without the items the conclusion rests on and a control without the same number of other items; the brief and question are unchanged, and removed items simply drop out of the index.
The two versions differ only in whether the grounds are present, so the source is held fixed.
We report the difference in closure rate between them (removed minus control; more negative means closure follows the evidence) and a two-class balanced accuracy per case, $\tfrac12[\mathbb{1}(\text{control closed}) + \mathbb{1}(\text{removed explicitly not closed})]$.

\emph{Conclusion and gap quality.}
A judge answers a fixed checklist about the final answer against the reference: whether the initiating cause is right, whether the causal chain matches, whether the answer overstates (presents as established what the evidence only suggests), what share of its causal claims cite evidence that was read and supports them, and whether it addresses alternative explanations.
Our summary metric is the share of answers whose cause is right \emph{and} not overstated.
For unclosed answers the judge also checks whether the gap names specific evidence, whether that evidence is truly absent from the case, and whether it matches the reference's reason for indeterminacy.
The main judge is GPT-6 Sol; because it is also the teacher (\S\ref{sec:dataset}), we repeat the checklist with Claude Sonnet 5 and Grok 4.7 (\S\ref{sec:quality}, Appendix~\ref{app:judges}).

\section{The \ours{} Dataset}
\label{sec:dataset}

\begin{figure*}[t]
\centering
\includegraphics[width=\textwidth]{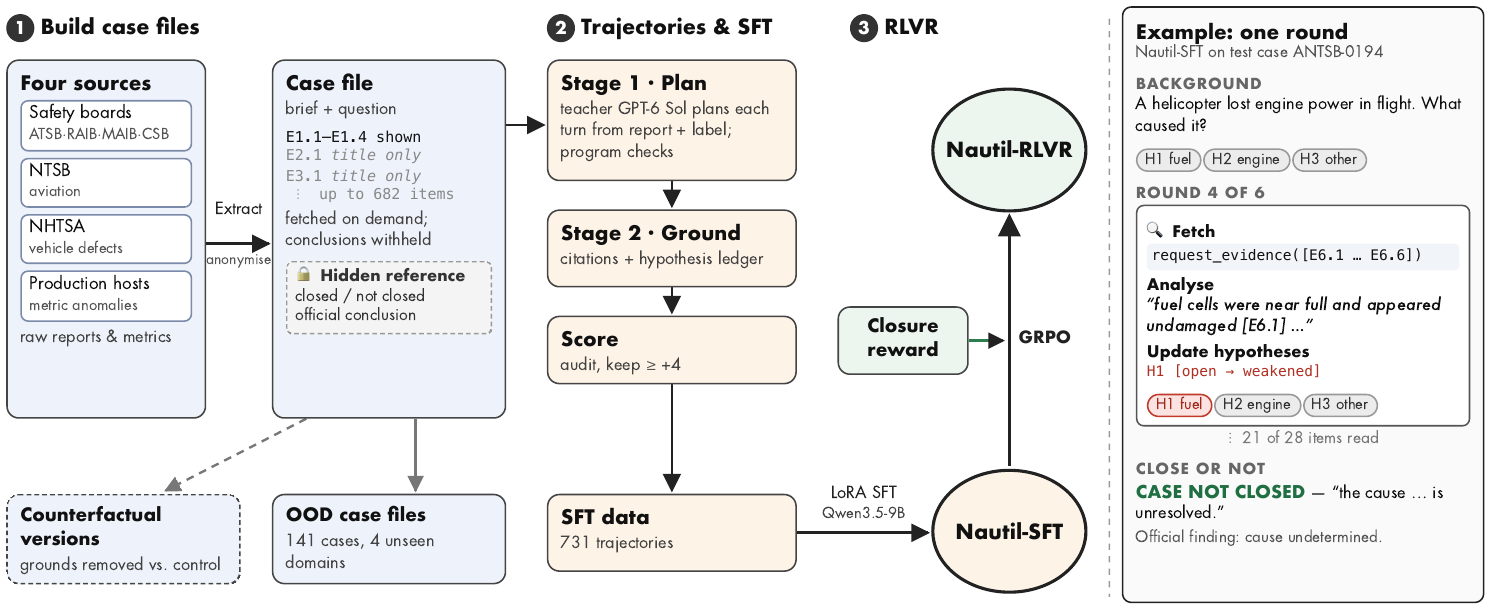}
\caption{Overview of \ours{}. (1) Case files are built from four sources: evidence items with neutral titles are shown on demand, conclusions are withheld, and a hidden reference records whether the case should close; counterfactual versions and the OOD set follow the same format. (2) The teacher plans each investigation from the report and its label, a program assembles and checks it, Stage 2 adds citations and the hypothesis ledger, and audited trajectories train \sft{}. (3) RLVR rewards only the closure decision, checked by a program. Right: one round of an actual \sft{} investigation on a test case whose official finding is undetermined; after its checks weaken the fuel and engine explanations, it leaves the case open.}
\label{fig:pipeline}
\vspace{-4pt}
\end{figure*}

\paragraph{Sources.}
\ours{} draws on four kinds of investigation (Table~\ref{tab:data}).
\emph{Boards}: reports of four transport and chemical-safety boards (ATSB, RAIB, MAIB, CSB).
\emph{NTSB}: aviation final reports 2014--2024.
\emph{NHTSA}: vehicle defect investigations, with opening and closing summaries.
\emph{Host}: anomaly windows from a year of production server monitoring at a large internet company.
Report cases take their label from whether the official report determined the cause. Host incidents have no report; their label is the teacher's closure decision (below), which we use for training and report only in the appendix.

\paragraph{Case files.}
For reports, an extraction pipeline (GPT-5.6 Sol) splits the text into factual, analytical and conclusion passages, turns factual passages into atomic evidence items with neutral titles, withholds any sentence that hints at the conclusion, writes the brief, anonymises identifiers, and records a hidden reference: whether the cause was determined, the official conclusion, rejected explanations and alternatives left open.
$E_1$ is fixed to the first four observation or measurement items of the primary source.
For hosts, selected metric series are rendered as charts and described in text by a vision-language model that sees each chart with its numbers table and may quote only those numbers; a program flags any number not in the table. Service names are replaced by neutral roles, and addresses, alert IDs and dates are removed.

\paragraph{Teacher trajectories.}
The teacher (GPT-6 Sol) sees the case file, the official analysis and the label, and plans an investigation: per turn, a note, the items to request and why, then a final answer (Stage 1).
A program assembles the dialogue with tool returns copied verbatim, rejecting plans that cite unrequested items, contradict the label or leak withheld identifiers.
Stage 2 adds exact citations and a ledger marking each hypothesis open, favoured or weakened, with every change citing visible evidence.
For hosts the teacher also sees the closing report of a run of the JustDiag diagnostic harness \citep{bi2026justdiag} on the incident.
Trajectories are retraced from the report's argument; their fetch order is not free exploration.

\paragraph{Audit.}
The teacher scores each trajectory from $-5$ to $+5$ on alignment with the source, evidence grounding, active fetching, hypothesis updates and honest closure; we keep scores $\ge 4$ without hard failures and allow one revision of the final answer for $+3$.
Of 1,345 generated report and host cases, 996 pass Stage 1, 818 pass Stage 2 and 518 (38.5\%) pass the audit; with supplements for host not-closed cases, a salvage round and the removal of 26 cases whose brief and evidence came from different incidents, the final set has 731 cases.
It is split 545 / 74 / 112 within each source, balancing labels and keeping each incident in one split.

\begin{table}[t]
\centering
\small
\caption{\ours{} by source. Items: median evidence items per case, over all cases of the source.}
\label{tab:data}
\setlength{\tabcolsep}{4.5pt}
\begin{tabular}{lrrrrc}
\toprule
Source & Candidates & Cases & Closed / not closed & Items & Train / Val / Test \\
\midrule
Boards (4 agencies) & 508 & 140 & 110 / 30 & 173 & 102 / 15 / 23 \\
NTSB (aviation) & 363 & 142 & 19 / 123 & 23.5 & 107 / 14 / 21 \\
NHTSA (vehicle defects) & 373 & 130 & 7 / 123 & 26 & 97 / 13 / 20 \\
Host (servers) & 1,713 & 319 & 259 / 60 & 145 & 239 / 32 / 48 \\
\midrule
Total & & 731 & 395 / 336 & 115 & 545 / 74 / 112 \\
\bottomrule
\end{tabular}
\vspace{-4pt}
\end{table}

\paragraph{Source and label.}
Table~\ref{tab:data} shows the structure that makes closure hard to measure: boards and hosts mostly close, NTSB and NHTSA mostly do not.
Answering by source alone gives 83.0 balanced accuracy on the 64 non-host test cases (23 should close, 41 should not).
This is why every closure result below is paired with the source-only rule and the within-source metric, and why the headline test set is non-host: host labels are teacher judgments, and the teacher itself, rerun under the same protocol, closes only 5 of the 39 host test cases labelled closed.

\paragraph{Out-of-distribution test set.}
141 cases from unseen domains: 83 reports (NTSB highway and pipeline, TSB Canada pipeline), 20 medical cases \citep{wu2025medcasereasoning} relabelled by whether the record contains the confirming finding, 18 microservice faults \citep{pham2025rcaeval}, and 20 Tennessee Eastman process runs \citep{downs1993plant,rieth2017additional}. Domain and label are again tied, so we report the OOD set per domain and as a secondary column.

\paragraph{Counterfactual versions.}
For the 62 test cases the teacher closes, the grounds are the items its final answer cites; versions remove 40\%, 75\% or 100\% of them, and the control removes as many other items. A leak check flags removed items that may be restated elsewhere; 20 cases have none.

\section{Training \ours{} Investigators}
\label{sec:training}

\paragraph{Supervised fine-tuning.}
We fine-tune Qwen3.5-9B \citep{qwen2026qwen35} with LoRA \citep{hu2022lora} (rank 16, $\alpha=32$, all linear layers) on the 545 training trajectories, with the native tool-call format, a 32k-token context, and loss on assistant turns only.
We train two epochs (546 steps) and select the epoch-2 checkpoint before touching the test set: validation loss flattens after it, validation closure accuracy is highest, and its samples disagree most on the closure decision, which leaves room for RL. We call it \sft{}.

\paragraph{RLVR with a closure-only reward.}
Starting from \sft{}, we run GRPO \citep{shao2024deepseekmath} in the same case environment. The reward is deliberately simple and fully checked by a program:
\begin{equation}
r = D + s, \qquad D = \begin{cases} +1 & \text{closure decision matches the label},\\ -1 & \text{it does not, or no closure marker}, \end{cases}
\label{eq:reward}
\end{equation}
where $s \le 0$ is a format penalty (no citation in the final answer, citing unread or non-existent items, turns beyond six, repeated requests, malformed calls, a missing marker).
Within each group, $s$ is centred separately among closing and non-closing rollouts, so that only $D$ separates the two decisions.
No judge is called during training.
Cases are drawn uniformly from the eight source $\times$ label strata, so that no source dominates the updates; host cases, whose labels are teacher judgments, make up a quarter of the draws.
Each step samples 22 cases with 8 rollouts each (temperature 1.0, up to 15 turns); we use learning rate $5\times10^{-6}$, asymmetric clipping (0.2 / 0.28) \citep{yu2025dapo}, no KL term, and 50 steps on two H100 GPUs. We select step 30 as \rlvr{} with a screening evaluation that included 30 test cases (22 of them non-host); Appendix~\ref{app:training} reports results with those cases removed.
We arrived at this reward after two judge-based variants performed worse (\S\ref{sec:components}); full settings are in Appendix~\ref{app:training}.

\section{Experiments}
\label{sec:experiments}

\paragraph{Setup.}
All investigators run the same protocol, prompt and tool.
We compare the untrained base model (Qwen3.5-9B), the base model prompted with ReAct \citep{yao2023react} or with our note-and-ledger format, Abstain-R1 \citep{zhai2026abstainr1} and its 3B base model, Gemini 3.8 Flash as a frontier model, \sft{}, \rlvr{}, and GPT-6 Sol as a teacher reference.
9B and 3B models are sampled three times per case; API models once, except Gemini on the non-host counterfactual cases (three times).
Unless noted, numbers are on the 64 non-host test cases, averaged over samples per case; intervals and host results are in Appendix~\ref{app:more}.

\subsection{Closure accuracy: are closure decisions right, and not by source?}
\label{sec:accuracy}

\begin{table}[t]
\centering
\small
\caption{Closure accuracy. Test: 64 non-host cases (23 should close, 41 should not). OOD: 141 cases in four unseen domains. Within-src.: balanced accuracy inside each report source, averaged; Boards: balanced accuracy inside boards, the only source with enough cases on both sides (18 / 5). The source-only rule scores 50 on both by construction.}
\label{tab:main}
\setlength{\tabcolsep}{4.5pt}
\begin{tabular}{lcccccc}
\toprule
 & \multicolumn{5}{c}{Test (non-host)} & OOD \\
\cmidrule(lr){2-6}\cmidrule(lr){7-7}
Model & Close$\to$cl. & Open$\to$op. & Balanced & Within-src. & Boards & Balanced \\
\midrule
Qwen3.5-9B (base) & 81.2 & 18.7 & 49.9 & 48.2 & 47.4 & 50.6 \\
\quad + ReAct prompt & 63.8 & 31.7 & 47.7 & 47.9 & 37.2 & 44.7 \\
\quad + ledger-format prompt & 92.8 & 17.9 & 55.3 & 56.9 & 52.0 & 56.9 \\
Qwen2.5-3B-Instruct & 10.1 & 68.3 & 39.2 & 40.9 & 33.7 & 31.9 \\
Abstain-R1 (3B) & 8.7 & 47.2 & 27.9 & 25.3 & 21.3 & 39.1 \\
Gemini 3.8 Flash$^{1}$ & 100.0 & 58.5 & 79.3 & 77.0 & 70.0 & 68.9 \\
\midrule
\sft{} & 44.9 & \textbf{93.5} & 69.2 & 60.4 & 70.2 & 61.7 \\
\rlvr{}$^{3}$ & 81.2 & 85.4 & \textbf{83.3} & 74.1 & \textbf{78.7} & 67.0 \\
\midrule
GPT-6 Sol (teacher)$^{1}$ & 78.3 & 85.4 & 81.8 & 73.0 & 71.7 & -- \\
Source-only rule & 78.3 & 87.8 & 83.0 & 50.0 & 50.0 & --$^{2}$ \\
\bottomrule
\end{tabular}

\vspace{2pt}
{\footnotesize $^{1}$One sample per case. $^{2}$Not applicable: OOD domains are unseen in training. $^{3}$Its checkpoint was chosen on a screening set that included 22 of these 64 test cases and 20 OOD cases (\S\ref{sec:training}); \sft{} was selected before any test-set use.}
\vspace{-4pt}
\end{table}

\emph{Yes: training turns a model that closes almost everything into one that separates the two kinds of case, and not only by source} (Table~\ref{tab:main}).
The base model closes 81\% of cases that should close but leaves only 18.7\% of undetermined cases open, for 49.9 balanced accuracy.
\sft{} leaves 93.5\% of undetermined cases open, the highest of all models, and reaches 69.2; \rlvr{} balances both sides (81.2 / 85.4) and reaches 83.3, on par with the teacher (81.8) and Gemini (79.3).
Gemini reaches its score by closing every case that should close and 41\% of those that should not.
Within each source, where the source-only rule is at 50, \rlvr{} reaches 74.1 (\sft{} 60.4, base 48.2), so its gains come from separating cases of the same source. Gemini's 77.0 is helped by NHTSA and NTSB having only 2 and 3 should-close cases; on boards, the only source with enough cases on both sides, \rlvr{} reaches 78.7, Gemini 70.0 and \sft{} 70.2 (differences within noise).
Over the whole test set, \rlvr{} ties the source-only rule (83.3 vs.\ 83.0), which is why the next two tests matter. The \sft{} numbers are unaffected by checkpoint selection; the \rlvr{} checkpoint was screened on a set that included 22 of the 64 test cases (Appendix~\ref{app:training}).

\subsection{Evidence dependence: do closures follow the evidence?}
\label{sec:dependence}

\begin{table}[t]
\centering
\small
\caption{Evidence dependence. Closure rate (\%) on the 23 non-host counterfactual cases: full case, control (non-grounding items removed), and all grounds removed. $\Delta$ = removed $-$ control (more negative = closure follows the evidence). Balanced: two-class balanced accuracy (control closed, removed explicitly not closed). Right: $\Delta$ on all 62 cases (incl.\ host) and on 20 cases without flagged leaks.}
\label{tab:cf}
\begin{tabular}{lcccccccc}
\toprule
 & \multicolumn{5}{c}{Non-host (23)} & All (62) & No leak (20) \\
\cmidrule(lr){2-6}\cmidrule(lr){7-7}\cmidrule(lr){8-8}
Model & Full & Control & Removed & $\Delta$ & Balanced & $\Delta$ & $\Delta$ \\
\midrule
Qwen3.5-9B (base) & 87.0 & 92.8 & 87.0 & $-$5.8 & 47.8 & $-$1.6 & $-$8.3 \\
Gemini 3.8 Flash & 97.1 & 100.0 & 92.8 & $-$7.2 & 53.6 & $-$4.3 & +6.7 \\
\midrule
\sft{} & 50.7 & 52.2 & 26.1 & \textbf{$-$26.1} & \textbf{62.3} & \textbf{$-$16.7} & \textbf{$-$23.3} \\
\rlvr{} & 75.4 & 76.8 & 60.9 & $-$15.9 & 58.0 & $-$9.1 & $-$5.0 \\
\midrule
GPT-6 Sol (teacher)$^{1}$ & 87.0 & 82.6 & 34.8 & $-$47.8 & 71.7 & $-$25.8 & $-$35.0 \\
\bottomrule
\end{tabular}

\vspace{2pt}
{\footnotesize Gemini: 3 samples per case on non-host cases and 1 on host cases, so its All and No-leak columns mix the two. $^{1}$One sample per case; the grounds are the items it cited as teacher, which favours it.}
\vspace{-4pt}
\end{table}

\begin{figure}[t]
\centering
\includegraphics[width=0.95\linewidth]{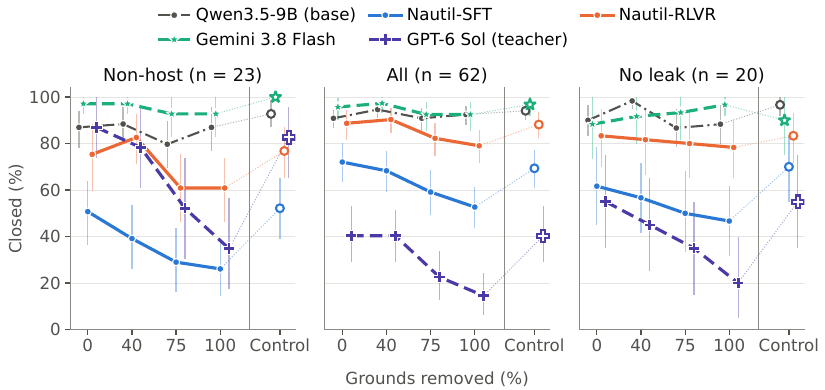}
\caption{Closure rate as more of a conclusion's grounds are removed (0--100\%), and on the control that removes as many non-grounding items (hollow markers). \sft{} closes less as its grounds disappear; the base model and Gemini keep closing. Dashed: API models (Gemini: 3 samples per case on non-host cases, 1 on host; GPT-6 Sol: 1). Bars: 95\% case-bootstrap intervals.}
\label{fig:cf}
\vspace{-4pt}
\end{figure}

\emph{Yes: after training, removing the grounds of a conclusion stops the model from closing} (Table~\ref{tab:cf}, Figure~\ref{fig:cf}).
With all grounds removed, \sft{} closes 26 points less than on the control, whose size is identical; the base model drops by 6 and Gemini by 7, still closing 87\% and 93\% of cases whose grounds are gone.
The effect grows with the share removed, holds on all 62 cases including hosts ($-$16.7 vs.\ $-$1.6 and $-$4.3), and on the 20 cases without any flagged leak ($-$23.3 vs.\ $-$8.3 and +6.7).
Because the two versions share source, brief and number of items, this test is designed to rule out both explanations that accuracy alone leaves open: recognising the source, and stopping just because evidence is scarce.
In two-class terms \sft{} scores 62.3 (base 47.8, Gemini 53.6); its difference from the base model is clear here and on all 62 cases, and from Gemini on the leak-free subset, while on the 23 non-host cases alone the intervals of $\Delta$ reach zero (Appendix~\ref{app:more}). As the grounds are those the teacher cited, a model might close a reduced case on other sufficient evidence; the control and the leak-free subset limit but do not exclude this.
\rlvr{} keeps a clear but smaller effect ($-$15.9), a trade-off we examine in \S\ref{sec:components}.
Appendix Figure~\ref{fig:caseC} shows a case where \sft{} closes on the control and stops when the recall record and crash test behind its conclusion are removed.

\subsection{Conclusion and gap quality: what does training change in the answers?}
\label{sec:quality}

\begin{table}[t]
\centering
\small
\caption{Conclusion and gap quality on the 64 non-host test cases (first sample). Left: GPT-6 Sol judge; right: Claude Sonnet 5 judge on the same answers. Overst.: presents as established what the evidence only suggests (lower is better). Right \& not ov.: cause right and not overstated. Supp.: share of causal claims citing read, supporting evidence. Alt.: addresses alternative explanations. Gap: gap score of unclosed answers (0--1).}
\label{tab:quality}
\footnotesize
\setlength{\tabcolsep}{3pt}
\begin{tabular}{lccccccc@{\hspace{10pt}}cc}
\toprule
 & \multicolumn{7}{c}{GPT-6 Sol judge} & \multicolumn{2}{c}{Sonnet 5 judge} \\
\cmidrule(lr){2-8}\cmidrule(lr){9-10}
Model & Score & Overst.$\downarrow$ & Cause & Right\,\&\,not ov. & Supp. & Alt. & Gap & Overst.$\downarrow$ & Right\,\&\,not ov. \\
\midrule
Qwen3.5-9B (base) & 0.30 & 97 & 54 & 3 & 26 & 19 & 0.80 & 81 & 10 \\
Gemini 3.8 Flash & 0.55 & 91 & 84 & 9 & 56 & 53 & 0.96 & 55 & 42 \\
\midrule
\sft{} & \textbf{0.58} & \textbf{35} & 56 & \textbf{43} & \textbf{78} & 38 & 0.38 & \textbf{11} & \textbf{63} \\
\rlvr{} & 0.53 & 38 & 62 & 36 & 76 & 17 & 0.39 & 20 & 50 \\
\midrule
GPT-6 Sol (teacher) & -- & -- & -- & -- & -- & -- & -- & 5 & 86 \\
\bottomrule
\end{tabular}
\vspace{-4pt}
\end{table}

\emph{Training changes restraint and grounding, not cause-finding} (Table~\ref{tab:quality}).
Overstatement falls from 97\% (base) to 35\% (\sft{}); the share of causal claims backed by evidence the model actually read rises from 26\% to 78\%; and answers that are both right about the cause and not overstated rise from 3\% to 43\%, against 9\% for Gemini.
The rate of naming the right cause barely moves (54\% to 56\%), and Gemini's is higher (84\%): the frontier model finds causes but states them beyond what the file supports.
The contrast is sharpest on officially undetermined cases: Gemini closes 17 of these 41 cases, and the judge, given an undetermined reference, rates all 17 conclusions as overstated, while \sft{} leaves 15 of the same 17 open.
\sft{} answers that make no causal claim at all are, without exception, \notclosed{} answers.
Overstatement rates depend on the judge (Sonnet 5: 11\% for \sft{}, 55\% for Gemini), but the ranking holds, and all three judges find trained models overstate far less (Appendix~\ref{app:judges}).
The limitation is on the other side of the decision: when \sft{} leaves a case open, its gap is often generic (gap score 0.38 against Gemini's 0.96), naming the specific missing evidence in 32\% of unclosed answers.

\subsection{Which components matter, and what does RL add?}
\label{sec:components}

\begin{table}[t]
\centering
\small
\caption{Components. SFT ablations are trained from the base model with the same two-epoch schedule; RL variants start from \sft{} and differ only in the reward. $\Delta_{\text{CF}}$: removed $-$ control on the 23 non-host counterfactual cases. Overst. and Score: GPT-6 Sol judge.}
\label{tab:ablation}
\footnotesize
\setlength{\tabcolsep}{3.2pt}
\begin{tabular}{lccccccc}
\toprule
Model & Close$\to$cl. & Open$\to$op. & Balanced & Within-src. & $\Delta_{\text{CF}}$ & Overst.$\downarrow$ & Score \\
\midrule
\sft{} & 44.9 & 93.5 & 69.2 & 60.4 & $-$26.1 & 35 & 0.58 \\
\quad w/o hypothesis ledger & 62.3 & 90.2 & 76.3 & 61.2 & $-$8.7 & 39 & 0.58 \\
\quad w/o not-closed trajectories & 94.2 & 0.0 & 47.1 & 48.8 & $-$8.7 & 84 & 0.41 \\
\quad 50\% of the data & 36.2 & 93.5 & 64.9 & 56.9 & -- & 36 & 0.56 \\
\quad 10\% of the data & 7.2 & 84.6 & 45.9 & 39.3 & -- & 31 & 0.58 \\
\midrule
RL: checklist always counted & 17.4 & 95.9 & 56.7 & 53.8 & $-$15.9 & 19 & 0.61 \\
RL: checklist if decision right & 56.5 & 89.4 & 73.0 & 63.0 & $-$20.3 & 33 & 0.60 \\
RL: decision only (\rlvr{}) & 81.2 & 85.4 & \textbf{83.3} & \textbf{74.1} & $-$15.9 & 38 & 0.53 \\
\bottomrule
\end{tabular}
\vspace{-4pt}
\end{table}

\emph{Not-closed demonstrations are the source of restraint.}
Without them, the model never leaves a case open (0.0\%) and overstates in 84\% of its answers (Table~\ref{tab:ablation}).
\emph{Without the hypothesis ledger, closure depends less on the evidence.}
Removing it leaves accuracy slightly higher but within noise of \sft{} (76.3 vs.\ 69.2), while the counterfactual effect shrinks from $-$26.1 to $-$8.7; the difference is clear on all 62 cases, and its interval reaches zero on the 23 non-host cases.
\emph{Data scale matters:} 10\%, 50\% and 100\% of the trajectories give 45.9, 64.9 and 69.2 (the number of steps grows with the data).

\emph{A closure-only reward gives the largest and cleanest RL gain.}
It raises balanced accuracy by 14.1 points and within-source accuracy from 60.4 to 74.1, removes missing closure markers (5.2\% to 0), and shortens investigations (5.5 to 4.6 turns); the OOD gain (5.3 points) is within noise.
Adding the judged checklist to the reward is harmful: counted on every rollout, it drives the policy towards never closing (17.4\% of should-close cases closed at step 10); counted only when the decision is right, it avoids the collapse but gains only 3.8 points.
A likely reason: when every rollout in a group is wrong, the checklist still ranks the wrong answers, and this spills over into reluctance to close.
The closure-only reward has a measurable cost: the counterfactual effect shrinks ($-$26.1 to $-$15.9; $-$23.3 to $-$5.0 without leaks), addressing alternatives drops (38\% to 17\%), and the conclusion score falls by 0.05: a trade-off between accuracy and evidence dependence, with \sft{} and \rlvr{} at its two ends.

\subsection{Further results}
\label{sec:further}

\paragraph{Prompting and single-turn abstention training do not substitute.}
ReAct prompting (47.7) and our ledger format as a prompt (55.3, with 95\% overstatement) leave the base model's closing habit intact (Table~\ref{tab:main}).
Abstain-R1 (27.9) and its base model (39.2) mostly fail to finish a multi-turn investigation (44\% and 32\% of runs end without a closure marker): abstention learned on single questions does not carry over to evidence gathering.

\paragraph{Out of distribution and long files.}
OOD balanced accuracy rises from 50.6 (base) to 61.7 (\sft{}) and 67.0 (\rlvr{}), close to Gemini (68.9). Because domain and label are tied in this set, this mostly reflects the direction of the decision; within individual domains no model is clearly above chance (Appendix~\ref{app:more}). On the 20 largest files, \rlvr{} closes 93\% of should-close cases when fetching its own evidence and 61\% when given the whole file, which in turn matches the main cause more often (Appendix~\ref{app:long}). Appendix Figure~\ref{fig:caseD} shows a failure in which \sft{} misreads a valve test.

\section{Limitations}
\label{sec:limitations}

\emph{Labels and sources.} The source of a case still predicts its label, so accuracy is paired with within-source and counterfactual tests; the within-source metric rests mostly on board cases. Host labels are teacher judgments the teacher does not reproduce; host results are in the appendix.
\emph{Judge, teacher and samples.} GPT-6 Sol is both teacher and main judge, so its investigator row is only a reference; two further judges agree on direction but not on absolute rates. API models are mostly sampled once; the non-host counterfactual set has 23 cases, and its grounds are defined by the teacher's citations. The \rlvr{} checkpoint was screened on 30 test cases, and a 33-case pilot set consulted while choosing the SFT epoch overlaps the OOD test set.
\emph{Data.} About 4\% of references have extraction errors, some anonymised words are damaged, and trajectories retrace the report rather than explore.
\emph{Open side of the decision.} When our models leave a case open they often do not name the specific missing evidence, where frontier models are far ahead; and \rlvr{} depends less on the evidence than \sft{}.

\section{Conclusion}
\label{sec:conclusion}

Closing a case only when the evidence supports it is a judgment that does not come with capability: a frontier model finds causes more often than ours and an untrained model about as often, yet both close cases their reports leave undetermined and overstate what the evidence shows.
Trained on audited investigation trajectories, a 9B model gains restraint and grounding: its closures move with the evidence, its claims cite what it read, and its conclusions are far less often overstated.
A closure-only RL reward lifts accuracy to the teacher's level at some cost in evidence dependence, whereas rewarding judged conclusion quality collapses the policy into never closing.
Because a case's source can stand in for its evidence, closure accuracy should be reported with a source-only baseline, within sources, and with evidence dependence.

\section*{AI use statement}
In this work, we used generative AI tools for literature search, discussion of research ideas, manuscript editing, and generation of experiment code.
We have reviewed all AI-assisted work and take responsibility for the final content of this work, including text, claims and artifacts produced with the aid of generative AI.

\section*{Reproducibility statement}
Our code repository (\url{https://github.com/etigerstudio/Nautil}) contains the code, prompts and training configurations of the pipeline, and no data: case extraction for reports, trajectory construction and auditing, splitting and counterfactual construction, the investigation environment, the investigator prompt and tool definition, and the SFT and RLVR code with its reward, judge prompts, configurations and unit tests.
The \ours{} dataset is available at \url{https://huggingface.co/datasets/etigerstudio/Nautil}, and the weights of \sft{} and \rlvr{} at \url{https://huggingface.co/etigerstudio/Nautil-SFT} and \url{https://huggingface.co/etigerstudio/Nautil-RLVR}. The production host cases and the code that builds them will be released after internal review.

\bibliography{nautil_related}

@inproceedings{rajpurkar2018know,
    title = "Know What You Don{'}t Know: Unanswerable Questions for {SQ}u{AD}",
    author = "Rajpurkar, Pranav  and
      Jia, Robin  and
      Liang, Percy",
    editor = "Gurevych, Iryna  and
      Miyao, Yusuke",
    booktitle = "Proceedings of the 56th Annual Meeting of the Association for Computational Linguistics (Volume 2: Short Papers)",
    month = jul,
    year = "2018",
    address = "Melbourne, Australia",
    publisher = "Association for Computational Linguistics",
    url = "https://aclanthology.org/P18-2124/",
    doi = "10.18653/v1/P18-2124",
    pages = "784--789"
}

@inproceedings{yin2023large,
    title = "Do Large Language Models Know What They Don{'}t Know?",
    author = "Yin, Zhangyue  and
      Sun, Qiushi  and
      Guo, Qipeng  and
      Wu, Jiawen  and
      Qiu, Xipeng  and
      Huang, Xuanjing",
    editor = "Rogers, Anna  and
      Boyd-Graber, Jordan  and
      Okazaki, Naoaki",
    booktitle = "Findings of the Association for Computational Linguistics: ACL 2023",
    month = jul,
    year = "2023",
    address = "Toronto, Canada",
    publisher = "Association for Computational Linguistics",
    url = "https://aclanthology.org/2023.findings-acl.551/",
    doi = "10.18653/v1/2023.findings-acl.551",
    pages = "8653--8665"
}

@inproceedings{zhang2024rtuning,
    title = "{R}-Tuning: Instructing Large Language Models to Say `{I} Don{'}t Know'",
    author = "Zhang, Hanning  and
      Diao, Shizhe  and
      Lin, Yong  and
      Fung, Yi  and
      Lian, Qing  and
      Wang, Xingyao  and
      Chen, Yangyi  and
      Ji, Heng  and
      Zhang, Tong",
    editor = "Duh, Kevin  and
      Gomez, Helena  and
      Bethard, Steven",
    booktitle = "Proceedings of the 2024 Conference of the North American Chapter of the Association for Computational Linguistics: Human Language Technologies (Volume 1: Long Papers)",
    month = jun,
    year = "2024",
    address = "Mexico City, Mexico",
    publisher = "Association for Computational Linguistics",
    url = "https://aclanthology.org/2024.naacl-long.394/",
    doi = "10.18653/v1/2024.naacl-long.394",
    pages = "7113--7139"
}

@inproceedings{zhai2026abstainr1,
    title = {{Abstain-R1: Calibrated Abstention and Post-Refusal Clarification via Verifiable {RL}}},
    author = "Zhai, Haotian  and
      Liang, Jingcheng  and
      Kang, Dongyeop",
    editor = "Liakata, Maria  and
      Moreira, Viviane P.  and
      Zhang, Jiajun  and
      Jurgens, David",
    booktitle = "Findings of the {A}ssociation for {C}omputational {L}inguistics: {ACL} 2026",
    month = jul,
    year = "2026",
    address = "San Diego, California, United States",
    publisher = "Association for Computational Linguistics",
    url = "https://aclanthology.org/2026.findings-acl.985/",
    doi = "10.18653/v1/2026.findings-acl.985",
    pages = "19674--19695",
    ISBN = "979-8-89176-395-1"
}

@inproceedings{kirichenko2026abstentionbench,
title={{AbstentionBench: Reasoning {LLM}s Fail on Unanswerable Questions}},
author={Polina Kirichenko and Mark Ibrahim and Kamalika Chaudhuri and Samuel J. Bell},
booktitle={The Thirty-ninth Annual Conference on Neural Information Processing Systems Datasets and Benchmarks Track},
year={2025},
url={https://openreview.net/forum?id=OkHC30LLpO}
}

@inproceedings{joren2025sufficient,
title={Sufficient Context: A New Lens on Retrieval Augmented Generation Systems},
author={Hailey Joren and Jianyi Zhang and Chun-Sung Ferng and Da-Cheng Juan and Ankur Taly and Cyrus Rashtchian},
booktitle={The Thirteenth International Conference on Learning Representations},
year={2025},
url={https://openreview.net/forum?id=Jjr2Odj8DJ}
}

@inproceedings{yang2024alignment,
title={Alignment for Honesty},
author={Yuqing Yang and Ethan Chern and Xipeng Qiu and Graham Neubig and Pengfei Liu},
booktitle={The Thirty-eighth Annual Conference on Neural Information Processing Systems},
year={2024},
url={https://openreview.net/forum?id=67K3Xlvw8L}
}

@inproceedings{wei2026truthrl,
title={Truth{RL}: Incentivizing Truthful {LLM}s via Reinforcement Learning},
author={Zhepei Wei and Xiao Yang and Kai Sun and Jiaqi Wang and Rulin Shao and Jingxiang Chen and Mohammad Kachuee and Teja Gollapudi and Yiwei Liao and Nicolas Scheffer and Rakesh Wanga and Anuj Kumar and Yu Meng and {Wen-tau} Yih and Xin Luna Dong},
booktitle={Forty-third International Conference on Machine Learning},
year={2026},
url={https://openreview.net/forum?id=TnJMJm3Fbi}
}

@inproceedings{damani2026beyond,
 author = {Damani, Mehul and Puri, Isha and Slocum, Stewart and Shenfeld, Idan and Choshen, Leshem and Kim, Yoon and Andreas, Jacob},
 booktitle = {International Conference on Learning Representations},
 editor = {C. Vondrick and B. Hariharan and C. Raffel and L. Pinto and D. Yang and A. Faust},
 pages = {59456--59493},
 title = {{Beyond Binary Rewards: Training LMs to Reason About Their Uncertainty}},
 url = {https://proceedings.iclr.cc/paper_files/paper/2026/file/615675cc6e94ddb1a783904fb178b5f6-Paper-Conference.pdf},
 volume = {2026},
 year = {2026}
}

@inproceedings{li2026questbench,
title={{QuestBench: Can {LLM}s ask the right question to acquire information in reasoning tasks?}},
author={Belinda Z. Li and Been Kim and Zi Wang},
booktitle={The Thirty-ninth Annual Conference on Neural Information Processing Systems Datasets and Benchmarks Track},
year={2025},
url={https://openreview.net/forum?id=gpwA9aZLTZ}
}

@inproceedings{li2024mediq,
 author = {Li, Shuyue Stella and Balachandran, Vidhisha and Feng, Shangbin and Ilgen, Jonathan S. and Pierson, Emma and Koh, Pang Wei and Tsvetkov, Yulia},
 booktitle = {Advances in Neural Information Processing Systems},
 doi = {10.52202/079017-0908},
 editor = {A. Globerson and L. Mackey and D. Belgrave and A. Fan and U. Paquet and J. Tomczak and C. Zhang},
 pages = {28858--28888},
 publisher = {Curran Associates, Inc.},
 title = {{MediQ: Question-Asking LLMs and a Benchmark for Reliable Interactive Clinical Reasoning}},
 url = {https://proceedings.neurips.cc/paper_files/paper/2024/file/32b80425554e081204e5988ab1c97e9a-Paper-Conference.pdf},
 volume = {37},
 year = {2024}
}

@misc{kadavath2022language,
      title={Language Models (Mostly) Know What They Know}, 
      author={Saurav Kadavath and Tom Conerly and Amanda Askell and Tom Henighan and Dawn Drain and Ethan Perez and Nicholas Schiefer and Zac Hatfield-Dodds and Nova DasSarma and Eli Tran-Johnson and Scott Johnston and Sheer El-Showk and Andy Jones and Nelson Elhage and Tristan Hume and Anna Chen and Yuntao Bai and Sam Bowman and Stanislav Fort and Deep Ganguli and Danny Hernandez and Josh Jacobson and Jackson Kernion and Shauna Kravec and Liane Lovitt and Kamal Ndousse and Catherine Olsson and Sam Ringer and Dario Amodei and Tom Brown and Jack Clark and Nicholas Joseph and Ben Mann and Sam McCandlish and Chris Olah and Jared Kaplan},
      year={2022},
      eprint={2207.05221},
      archivePrefix={arXiv},
      primaryClass={cs.CL},
      url={https://arxiv.org/abs/2207.05221}, 
}

@misc{kalai2025why,
      title={Why Language Models Hallucinate}, 
      author={Adam Tauman Kalai and Ofir Nachum and Santosh S. Vempala and Edwin Zhang},
      year={2025},
      eprint={2509.04664},
      archivePrefix={arXiv},
      primaryClass={cs.CL},
      url={https://arxiv.org/abs/2509.04664}, 
}

@misc{handler2026quantifying,
      title={{Quantifying and Mitigating Premature Closure in Frontier LLMs}}, 
      author={Rebecca Handler and Suhana Bedi and Nigam Shah},
      year={2026},
      eprint={2605.15000},
      archivePrefix={arXiv},
      primaryClass={cs.CL},
      url={https://arxiv.org/abs/2605.15000}, 
}

@misc{che2026abstention,
      title={{Abstention as an Action Can Kill Both the Reward Gradient and the KL Anchor: Collapse Law and Repair for Error-Penalized Reinforcement Learning}}, 
      author={Xujun Che and Yuchen Yuan and Weida Zhao and Chenyang Yu},
      year={2026},
      eprint={2608.00301},
      archivePrefix={arXiv},
      primaryClass={cs.LG},
      url={https://arxiv.org/abs/2608.00301}, 
}

@misc{liu2025cart,
      title={{CaRT: Teaching LLM Agents to Know When They Know Enough}}, 
      author={Grace Liu and Yuxiao Qu and Jeff Schneider and Aarti Singh and Aviral Kumar},
      year={2025},
      eprint={2510.08517},
      archivePrefix={arXiv},
      primaryClass={cs.AI},
      url={https://arxiv.org/abs/2510.08517}, 
}

@misc{luo2026agentic,
      title={Agentic Abstention: Do Agents Know When to Stop Instead of Act?}, 
      author={Han Luo and Bingbing Wen and Lucy Lu Wang},
      year={2026},
      eprint={2606.28733},
      archivePrefix={arXiv},
      primaryClass={cs.AI},
      url={https://arxiv.org/abs/2606.28733}, 
}

@misc{liu2026agentabstain,
      title={{AgentAbstain: Do LLM Agents Know When Not to Act?}}, 
      author={Xun Liu and Yi Evie Zhang and Vira Kasprova and Parisa Rabbani and Pardis Sadat Zahraei and Tianyu Zhang and Ali Ebrahimpour-Boroojeny and Varun Chandrasekaran},
      year={2026},
      eprint={2607.10059},
      archivePrefix={arXiv},
      primaryClass={cs.AI},
      url={https://arxiv.org/abs/2607.10059}, 
}

@inproceedings{song2025hallucination,
    title = "The Hallucination Tax of Reinforcement Finetuning",
    author = "Song, Linxin  and
      Shi, Taiwei  and
      Zhao, Jieyu",
    editor = "Christodoulopoulos, Christos  and
      Chakraborty, Tanmoy  and
      Rose, Carolyn  and
      Peng, Violet",
    booktitle = "Findings of the Association for Computational Linguistics: EMNLP 2025",
    month = nov,
    year = "2025",
    address = "Suzhou, China",
    publisher = "Association for Computational Linguistics",
    url = "https://aclanthology.org/2025.findings-emnlp.112/",
    doi = "10.18653/v1/2025.findings-emnlp.112",
    pages = "2105--2120",
    ISBN = "979-8-89176-335-7"
}

@inproceedings{yao2023react,
title={{ReAct: Synergizing Reasoning and Acting in Language Models}},
author={Shunyu Yao and Jeffrey Zhao and Dian Yu and Nan Du and Izhak Shafran and Karthik R Narasimhan and Yuan Cao},
booktitle={The Eleventh International Conference on Learning Representations },
year={2023},
url={https://openreview.net/forum?id=WE_vluYUL-X}
}

@inproceedings{jin2025searchr1,
title={{Search-R1: Training {LLM}s to Reason and Leverage Search Engines with Reinforcement Learning}},
author={Bowen Jin and Hansi Zeng and Zhenrui Yue and Jinsung Yoon and Sercan O Arik and Dong Wang and Hamed Zamani and Jiawei Han},
booktitle={Second Conference on Language Modeling},
year={2025},
url={https://openreview.net/forum?id=Rwhi91ideu}
}

@misc{song2025r1searcher,
      title={{R1-Searcher: Incentivizing the Search Capability in LLMs via Reinforcement Learning}},
      author={Huatong Song and Jinhao Jiang and Yingqian Min and Jie Chen and Zhipeng Chen and Wayne Xin Zhao and Lei Fang and Ji-Rong Wen},
      year={2025},
      eprint={2503.05592},
      archivePrefix={arXiv},
      primaryClass={cs.AI},
      url={https://arxiv.org/abs/2503.05592},
}

@misc{wang2025ragen,
      title={{RAGEN: Understanding Self-Evolution in LLM Agents via Multi-Turn Reinforcement Learning}},
      author={Zihan Wang and Kangrui Wang and Qineng Wang and Pingyue Zhang and Linjie Li and Zhengyuan Yang and Xing Jin and Kefan Yu and Minh Nhat Nguyen and Licheng Liu and Eli Gottlieb and Yiping Lu and Kyunghyun Cho and Jiajun Wu and Li Fei-Fei and Lijuan Wang and Yejin Choi and Manling Li},
      year={2025},
      eprint={2504.20073},
      archivePrefix={arXiv},
      primaryClass={cs.LG},
      url={https://arxiv.org/abs/2504.20073},
}

@misc{nori2025sequential,
      title={Sequential Diagnosis with Language Models},
      author={Harsha Nori and Mayank Daswani and Christopher Kelly and Scott Lundberg and Marco Tulio Ribeiro and Marc Wilson and Xiaoxuan Liu and Viknesh Sounderajah and Jonathan Carlson and Matthew P Lungren and Bay Gross and Peter Hames and Mustafa Suleyman and Dominic King and Eric Horvitz},
      year={2025},
      eprint={2506.22405},
      archivePrefix={arXiv},
      primaryClass={cs.CL},
      url={https://arxiv.org/abs/2506.22405},
}

@misc{marozzo2026llm,
      title={{LLM-as-an-Investigator: Evidence-First Reasoning for Robust Interactive Problem Diagnosis}},
      author={Fabrizio Marozzo and Pietro Liò},
      year={2026},
      eprint={2606.13220},
      archivePrefix={arXiv},
      primaryClass={cs.AI},
      url={https://arxiv.org/abs/2606.13220},
}

@misc{chen2026holmes,
      title={{Holmes: An Evidence-Grounded LLM Agent for Auditable DDoS Investigation in Cloud Networks}},
      author={Haodong Chen and Ziheng Zhang and Jinghui Jiang and Qiang Su and Qiao Xiang},
      year={2026},
      eprint={2601.14601},
      archivePrefix={arXiv},
      primaryClass={cs.CR},
      url={https://arxiv.org/abs/2601.14601},
}

@article{guo2025deepseekr1,
  title={{DeepSeek-R1 incentivizes reasoning in LLMs through reinforcement learning}},
  volume={645},
  ISSN={1476-4687},
  url={http://dx.doi.org/10.1038/s41586-025-09422-z},
  DOI={10.1038/s41586-025-09422-z},
  number={8081},
  journal={Nature},
  publisher={Springer Science and Business Media LLC},
  author={Guo, Daya and Yang, Dejian and Zhang, Haowei and Song, Junxiao and Wang, Peiyi and Zhu, Qihao and Xu, Runxin and Zhang, Ruoyu and Ma, Shirong and Bi, Xiao and Zhang, Xiaokang and Yu, Xingkai and Wu, Yu and Wu, Z. F. and Gou, Zhibin and Shao, Zhihong and Li, Zhuoshu and Gao, Ziyi and Liu, Aixin and Xue, Bing and Wang, Bingxuan and Wu, Bochao and Feng, Bei and Lu, Chengda and Zhao, Chenggang and Deng, Chengqi and Ruan, Chong and Dai, Damai and Chen, Deli and Ji, Dongjie and Li, Erhang and Lin, Fangyun and Dai, Fucong and Luo, Fuli and Hao, Guangbo and Chen, Guanting and Li, Guowei and Zhang, H. and Xu, Hanwei and Ding, Honghui and Gao, Huazuo and Qu, Hui and Li, Hui and Guo, Jianzhong and Li, Jiashi and Chen, Jingchang and Yuan, Jingyang and Tu, Jinhao and Qiu, Junjie and Li, Junlong and Cai, J. L. and Ni, Jiaqi and Liang, Jian and Chen, Jin and Dong, Kai and Hu, Kai and You, Kaichao and Gao, Kaige and Guan, Kang and Huang, Kexin and Yu, Kuai and Wang, Lean and Zhang, Lecong and Zhao, Liang and Wang, Litong and Zhang, Liyue and Xu, Lei and Xia, Leyi and Zhang, Mingchuan and Zhang, Minghua and Tang, Minghui and Zhou, Mingxu and Li, Meng and Wang, Miaojun and Li, Mingming and Tian, Ning and Huang, Panpan and Zhang, Peng and Wang, Qiancheng and Chen, Qinyu and Du, Qiushi and Ge, Ruiqi and Zhang, Ruisong and Pan, Ruizhe and Wang, Runji and Chen, R. J. and Jin, R. L. and Chen, Ruyi and Lu, Shanghao and Zhou, Shangyan and Chen, Shanhuang and Ye, Shengfeng and Wang, Shiyu and Yu, Shuiping and Zhou, Shunfeng and Pan, Shuting and Li, S. S. and Zhou, Shuang and Wu, Shaoqing and Yun, Tao and Pei, Tian and Sun, Tianyu and Wang, T. and Zeng, Wangding and Liu, Wen and Liang, Wenfeng and Gao, Wenjun and Yu, Wenqin and Zhang, Wentao and Xiao, W. L. and An, Wei and Liu, Xiaodong and Wang, Xiaohan and Chen, Xiaokang and Nie, Xiaotao and Cheng, Xin and Liu, Xin and Xie, Xin and Liu, Xingchao and Yang, Xinyu and Li, Xinyuan and Su, Xuecheng and Lin, Xuheng and Li, X. Q. and Jin, Xiangyue and Shen, Xiaojin and Chen, Xiaosha and Sun, Xiaowen and Wang, Xiaoxiang and Song, Xinnan and Zhou, Xinyi and Wang, Xianzu and Shan, Xinxia and Li, Y. K. and Wang, Y. Q. and Wei, Y. X. and Zhang, Yang and Xu, Yanhong and Li, Yao and Zhao, Yao and Sun, Yaofeng and Wang, Yaohui and Yu, Yi and Zhang, Yichao and Shi, Yifan and Xiong, Yiliang and He, Ying and Piao, Yishi and Wang, Yisong and Tan, Yixuan and Ma, Yiyang and Liu, Yiyuan and Guo, Yongqiang and Ou, Yuan and Wang, Yuduan and Gong, Yue and Zou, Yuheng and He, Yujia and Xiong, Yunfan and Luo, Yuxiang and You, Yuxiang and Liu, Yuxuan and Zhou, Yuyang and Zhu, Y. X. and Huang, Yanping and Li, Yaohui and Zheng, Yi and Zhu, Yuchen and Ma, Yunxian and Tang, Ying and Zha, Yukun and Yan, Yuting and Ren, Z. Z. and Ren, Zehui and Sha, Zhangli and Fu, Zhe and Xu, Zhean and Xie, Zhenda and Zhang, Zhengyan and Hao, Zhewen and Ma, Zhicheng and Yan, Zhigang and Wu, Zhiyu and Gu, Zihui and Zhu, Zijia and Liu, Zijun and Li, Zilin and Xie, Ziwei and Song, Ziyang and Pan, Zizheng and Huang, Zhen and Xu, Zhipeng and Zhang, Zhongyu and Zhang, Zhen},
  year={2025},
  month=sep,
  pages={633–638}
}

@misc{shao2024deepseekmath,
      title={{DeepSeekMath: Pushing the Limits of Mathematical Reasoning in Open Language Models}}, 
      author={Zhihong Shao and Peiyi Wang and Qihao Zhu and Runxin Xu and Junxiao Song and Xiao Bi and Haowei Zhang and Mingchuan Zhang and Y. K. Li and Y. Wu and Daya Guo},
      year={2024},
      eprint={2402.03300},
      archivePrefix={arXiv},
      primaryClass={cs.CL},
      url={https://arxiv.org/abs/2402.03300}, 
}

@inproceedings{gunjal2026rubrics,
 author = {Gunjal, Anisha and Wang, Anthony and Lau, Elaine and Nath, Vaskar and He, Yunzhong and Liu, Bing and Hendryx, Sean},
 booktitle = {International Conference on Learning Representations},
 editor = {C. Vondrick and B. Hariharan and C. Raffel and L. Pinto and D. Yang and A. Faust},
 pages = {127924--127945},
 title = {Rubrics as Rewards: Reinforcement Learning Beyond Verifiable Domains},
 url = {https://proceedings.iclr.cc/paper_files/paper/2026/file/cfd7bee7a651ee9af525098ef67a9e45-Paper-Conference.pdf},
 volume = {2026},
 year = {2026}
}

@inproceedings{skalse2022defining,
 author = {Skalse, Joar and Howe, Nikolaus and Krasheninnikov, Dmitrii and Krueger, David},
 booktitle = {Advances in Neural Information Processing Systems},
 doi = {10.52202/068431-0687},
 editor = {S. Koyejo and S. Mohamed and A. Agarwal and D. Belgrave and K. Cho and A. Oh},
 pages = {9460--9471},
 publisher = {Curran Associates, Inc.},
 title = {Defining and Characterizing Reward Gaming},
 url = {https://proceedings.neurips.cc/paper_files/paper/2022/file/3d719fee332caa23d5038b8a90e81796-Paper-Conference.pdf},
 volume = {35},
 year = {2022}
}

@InProceedings{gao2023scaling,
  title = 	 {Scaling Laws for Reward Model Overoptimization},
  author =       {Gao, Leo and Schulman, John and Hilton, Jacob},
  booktitle = 	 {Proceedings of the 40th International Conference on Machine Learning},
  pages = 	 {10835--10866},
  year = 	 {2023},
  editor = 	 {Krause, Andreas and Brunskill, Emma and Cho, Kyunghyun and Engelhardt, Barbara and Sabato, Sivan and Scarlett, Jonathan},
  volume = 	 {202},
  series = 	 {Proceedings of Machine Learning Research},
  month = 	 {23--29 Jul},
  publisher =    {PMLR},
  url = 	 {https://proceedings.mlr.press/v202/gao23h.html}
}

@inproceedings{zheng2023judging,
 author = {Zheng, Lianmin and Chiang, Wei-Lin and Sheng, Ying and Zhuang, Siyuan and Wu, Zhanghao and Zhuang, Yonghao and Lin, Zi and Li, Zhuohan and Li, Dacheng and Xing, Eric and Zhang, Hao and Gonzalez, Joseph and Stoica, Ion},
 booktitle = {Advances in Neural Information Processing Systems},
 doi = {10.52202/075280-2020},
 editor = {A. Oh and T. Naumann and A. Globerson and K. Saenko and M. Hardt and S. Levine},
 pages = {46595--46623},
 publisher = {Curran Associates, Inc.},
 title = {{Judging LLM-as-a-Judge with MT-Bench and Chatbot Arena}},
 url = {https://proceedings.neurips.cc/paper_files/paper/2023/file/91f18a1287b398d378ef22505bf41832-Paper-Datasets_and_Benchmarks.pdf},
 volume = {36},
 year = {2023}
}

@inproceedings{panickssery2024llm,
 author = {Panickssery, Arjun and Bowman, Samuel R. and Feng, Shi},
 booktitle = {Advances in Neural Information Processing Systems},
 doi = {10.52202/079017-2197},
 editor = {A. Globerson and L. Mackey and D. Belgrave and A. Fan and U. Paquet and J. Tomczak and C. Zhang},
 pages = {68772--68802},
 publisher = {Curran Associates, Inc.},
 title = {{LLM Evaluators Recognize and Favor Their Own Generations}},
 url = {https://proceedings.neurips.cc/paper_files/paper/2024/file/7f1f0218e45f5414c79c0679633e47bc-Paper-Conference.pdf},
 volume = {37},
 year = {2024}
}

@article{geirhos2020shortcut,
  title={Shortcut learning in deep neural networks},
  volume={2},
  ISSN={2522-5839},
  url={http://dx.doi.org/10.1038/s42256-020-00257-z},
  DOI={10.1038/s42256-020-00257-z},
  number={11},
  journal={Nature Machine Intelligence},
  publisher={Springer Science and Business Media LLC},
  author={Geirhos, Robert and Jacobsen, Jörn-Henrik and Michaelis, Claudio and Zemel, Richard and Brendel, Wieland and Bethge, Matthias and Wichmann, Felix A.},
  year={2020},
  month=Nov,
  pages={665–673}
}

@inproceedings{gururangan2018annotation,
    title = "Annotation Artifacts in Natural Language Inference Data",
    author = "Gururangan, Suchin  and
      Swayamdipta, Swabha  and
      Levy, Omer  and
      Schwartz, Roy  and
      Bowman, Samuel R.  and
      Smith, Noah A.",
    editor = "Walker, Marilyn  and
      Ji, Heng  and
      Stent, Amanda",
    booktitle = "Proceedings of the 2018 Conference of the North {A}merican Chapter of the Association for Computational Linguistics: Human Language Technologies, Volume 2 (Short Papers)",
    month = jun,
    year = "2018",
    address = "New Orleans, Louisiana",
    publisher = "Association for Computational Linguistics",
    url = "https://aclanthology.org/N18-2017/",
    doi = "10.18653/v1/N18-2017",
    pages = "107--112"
}

@inproceedings{gardner2020evaluating,
    title = "Evaluating Models' Local Decision Boundaries via Contrast Sets",
    author = "Gardner, Matt  and
      Artzi, Yoav  and
      Basmov, Victoria  and
      Berant, Jonathan  and
      Bogin, Ben  and
      Chen, Sihao  and
      Dasigi, Pradeep  and
      Dua, Dheeru  and
      Elazar, Yanai  and
      Gottumukkala, Ananth  and
      Gupta, Nitish  and
      Hajishirzi, Hannaneh  and
      Ilharco, Gabriel  and
      Khashabi, Daniel  and
      Lin, Kevin  and
      Liu, Jiangming  and
      Liu, Nelson F.  and
      Mulcaire, Phoebe  and
      Ning, Qiang  and
      Singh, Sameer  and
      Smith, Noah A.  and
      Subramanian, Sanjay  and
      Tsarfaty, Reut  and
      Wallace, Eric  and
      Zhang, Ally  and
      Zhou, Ben",
    editor = "Cohn, Trevor  and
      He, Yulan  and
      Liu, Yang",
    booktitle = "Findings of the Association for Computational Linguistics: EMNLP 2020",
    month = nov,
    year = "2020",
    address = "Online",
    publisher = "Association for Computational Linguistics",
    url = "https://aclanthology.org/2020.findings-emnlp.117/",
    doi = "10.18653/v1/2020.findings-emnlp.117",
    pages = "1307--1323"
}

@inproceedings{kaushik2020learning,
      title={Learning the Difference that Makes a Difference with Counterfactually-Augmented Data},
      author={Divyansh Kaushik and Eduard Hovy and Zachary C. Lipton},
      booktitle={International Conference on Learning Representations},
      year={2020},
      url={https://openreview.net/forum?id=Sklgs0NFvr},
}

@misc{qiu2026evidence,
      title={{Evidence Absence Is Not Evidence Insufficiency: Diagnosing NEI Construction Artifacts in Fact Verification}}, 
      author={Jingxi Qiu and Zeyu Han and Cheng Huang},
      year={2026},
      eprint={2605.26663},
      archivePrefix={arXiv},
      primaryClass={cs.CL},
      url={https://arxiv.org/abs/2605.26663}, 
}

@inproceedings{wang2024rcagent, series={CIKM ’24}, title={{RCAgent: Cloud Root Cause Analysis by Autonomous Agents with Tool-Augmented Large Language Models}}, url={http://dx.doi.org/10.1145/3627673.3680016}, DOI={10.1145/3627673.3680016}, booktitle={Proceedings of the 33rd ACM International Conference on Information and Knowledge Management}, publisher={ACM}, author={Wang, Zefan and Liu, Zichuan and Zhang, Yingying and Zhong, Aoxiao and Wang, Jihong and Yin, Fengbin and Fan, Lunting and Wu, Lingfei and Wen, Qingsong}, year={2024}, month=Oct, pages={4966–4974}, collection={CIKM ’24} }

@inproceedings{xu2025openrca,
 author = {Xu, Junjielong and Zhang, Qinan and Zhong, Zhiqing and He, Shilin and Zhang, Chaoyun and Lin, Qingwei and Pei, Dan and He, Pinjia and Zhang, Dongmei and Zhang, Qi},
 booktitle = {International Conference on Learning Representations},
 editor = {Y. Yue and A. Garg and N. Peng and F. Sha and R. Yu},
 pages = {84562--84590},
 title = {{OpenRCA: Can Large Language Models Locate the Root Cause of Software Failures?}},
 url = {https://proceedings.iclr.cc/paper_files/paper/2025/file/d29b8d53678015079e1d245c023e49d2-Paper-Conference.pdf},
 volume = {2025},
 year = {2025}
}

@inproceedings{pham2025rcaeval, series={WWW ’25}, title={{RCAEval: A Benchmark for Root Cause Analysis of Microservice Systems with Telemetry Data}}, url={http://dx.doi.org/10.1145/3701716.3715290}, DOI={10.1145/3701716.3715290}, booktitle={Companion Proceedings of the ACM on Web Conference 2025}, publisher={ACM}, author={Pham, Luan and Zhang, Hongyu and Ha, Huong and Salim, Flora and Zhang, Xiuzhen}, year={2025}, month=May, pages={777–780}, collection={WWW ’25} }

@inproceedings{chen2025aiopslab,
 author = {Chen, Yinfang and Shetty, Manish and Somashekar, Gagan and Ma, Minghua and Simmhan, Yogesh and Mace, Jonathan and Bansal, Chetan and Wang, Rujia and Rajmohan, Saravan},
 booktitle = {Proceedings of Machine Learning and Systems},
 editor = {M. Zaharia and G. Joshi and Y. Lin},
 pages = {},
 publisher = {MLSys},
 title = {{AIOpsLab: A Holistic Framework to Evaluate AI Agents for Enabling Autonomous Clouds}},
 url = {https://proceedings.mlsys.org/paper_files/paper/2025/file/d1f9e4a9f109b6e8b75ed362736f22ec-Paper-Conference.pdf},
 volume = {7},
 year = {2025}
}

@misc{wu2025medcasereasoning,
      title={{MedCaseReasoning: Evaluating and learning diagnostic reasoning from clinical case reports}}, 
      author={Kevin Wu and Eric Wu and Rahul Thapa and Kevin Wei and Angela Zhang and Arvind Suresh and Jacqueline J. Tao and Min Woo Sun and Alejandro Lozano and James Zou},
      year={2025},
      eprint={2505.11733},
      archivePrefix={arXiv},
      primaryClass={cs.CL},
      url={https://arxiv.org/abs/2505.11733}, 
}

@article{downs1993plant, title={A plant-wide industrial process control problem}, volume={17}, ISSN={0098-1354}, url={http://dx.doi.org/10.1016/0098-1354(93)80018-I}, DOI={10.1016/0098-1354(93)80018-i}, number={3}, journal={Computers \& Chemical Engineering}, publisher={Elsevier BV}, author={Downs, J.J. and Vogel, E.F.}, year={1993}, month=Mar, pages={245–255} }

@misc{rieth2017additional,
  doi = {10.7910/DVN/6C3JR1},
  url = {https://dataverse.harvard.edu/citation?persistentId=doi:10.7910/DVN/6C3JR1},
  author = {Rieth, Cory A. and Amsel, Ben D. and Tran, Randy and Cook, Maia B.},
  title = {{Additional Tennessee Eastman Process Simulation Data for Anomaly Detection Evaluation}},
  publisher = {Harvard Dataverse},
  year = {2017}
}

@misc{bi2026justdiag,
      title={{JustDiag!}: A Diagnostic Justification Engine for Accountable Root Cause Analysis},
      author={Tingzhu Bi and Xinrui Jiang and Xun Zhang and Pengcheng Su and Congjie He and Jinglin Li and Ping Wang and Meng Ma},
      year={2026},
      eprint={2606.19407},
      archivePrefix={arXiv},
      primaryClass={cs.SE},
      url={https://arxiv.org/abs/2606.19407},
}

@misc{qwen2026qwen35,
    title  = {{Qwen3.5}: Towards Native Multimodal Agents},
    author = {{Qwen Team}},
    year   = {2026},
    month  = {February},
    url    = {https://qwen.ai/blog?id=qwen3.5}
}

@inproceedings{hu2022lora,
title={Lo{RA}: Low-Rank Adaptation of Large Language Models},
author={Edward J. Hu and Yelong Shen and Phillip Wallis and Zeyuan Allen-Zhu and Yuanzhi Li and Shean Wang and Lu Wang and Weizhu Chen},
booktitle={International Conference on Learning Representations},
year={2022},
url={https://openreview.net/forum?id=nZeVKeeFYf9}
}

@inproceedings{yu2025dapo,
 author = {Yu, Qiying and Zhang, Zheng and Zhu, Ruofei and Yuan, Yufeng and Zuo, Xiaochen and Yue, Yu and Dai, Weinan and Fan, Tiantian and Liu, Gaohong and Liu, Juncai and Liu, LingJun and Liu, Xin and Lin, Haibin and Lin, Zhiqi and Ma, Bole and Sheng, Guangming and Tong, Yuxuan and Zhang, Chi and Zhang, Mofan and Zhang, Ru and Zhang, Wang and Zhu, Hang and Zhu, Jinhua and Chen, Jiaze and Chen, Jiangjie and Wang, Chengyi and Yu, Hongli and Song, Yuxuan and Wei, Xiangpeng and Zhou, Hao and Liu, Jingjing and Ma, Wei-Ying and Zhang, Ya-Qin and Yan, Lin and Wu, Yonghui and Wang, Mingxuan},
 booktitle = {Advances in Neural Information Processing Systems},
 doi = {10.52202/085713-3775},
 editor = {D. Belgrave and C. Zhang and H. Lin and R. Pascanu and P. Koniusz and M. Ghassemi and N. Chen},
 pages = {113222--113244},
 publisher = {Curran Associates, Inc.},
 title = {{DAPO}: An Open-Source {LLM} Reinforcement Learning System at Scale},
 url = {https://proceedings.neurips.cc/paper_files/paper/2025/file/a4277440d50f1f15d2cb4c14f7e0c0d2-Paper-Conference.pdf},
 volume = {38},
 year = {2025}
}
\bibliographystyle{iclr2027_conference}

\appendix

\section{Dataset construction}
\label{app:data}

\paragraph{Sources and licences.}
Candidate pools: boards 508 reports (ATSB 86, RAIB 158, MAIB 170, CSB 94), NTSB 363 aviation final reports (2014--2024), NHTSA 373 defect investigations, and 1,713 host anomaly windows.
ATSB reports are CC BY 4.0, RAIB and MAIB reports are under the Open Government Licence v3, and CSB, NTSB and NHTSA documents are US government works.
Host windows were flagged by a rolling median/MAD detector, so they are anomalies, not verified root-cause labels.
Medical case reports were not used for training because their licences are mixed; they appear only in the OOD test set.

\paragraph{Report extraction.}
The extractor (GPT-5.6 Sol) runs seven steps: (1) label each passage as factual, analytical or conclusion; (2) split factual passages into atomic evidence items with neutral titles, moving any sentence that hints at a conclusion to a withheld list; (3) build the source table; (4) write the brief; (5) anonymise; (6) write the hidden reference (determined or not, closure object, official conclusion verbatim, rejected explanations, alternatives left open); (7) review.
$E_1$ is regrouped to the first four observation or measurement items of the primary source, so that the opening record does not depend on document layout.

\paragraph{Host textualisation.}
For each window we select metrics, render one chart per six metrics (baseline median, spread band, largest departure), describe each chart with a vision-language model that also receives the numbers table, and assemble the case with the alert record as $E_1$. A program flags any number in a description that is not in the table.

\paragraph{Anonymisation.}
For reports, a model lists identifiers (dates, locations, operators, registrations, report numbers) and a program replaces them in two passes, followed by a forced sweep. The replacement matched substrings rather than whole words, which damaged some ordinary words (e.g.\ ``thunder'' became ``thundethe'') in at least 63 cases. We did not alter generated trajectories; items never fetched by the teacher were re-anonymised with whole-word matching (942 items). For hosts, service names become neutral roles; IP addresses, alert IDs, links and calendar dates are removed.

\paragraph{Host labels.}
The JustDiag diagnostic harness \citep{bi2026justdiag} (with GLM-5.3-Flash) was run on every host window; runs were scored by GPT-5.6 Sol and those scoring $\ge +1$ entered the pool. When writing a host trajectory, the teacher sees this run's closing report as a reference, so the closure marker of a host trajectory is the teacher's judgment, not a ground truth. Not-closed host examples come from 79 unfinished diagnosis runs, after removing hosts that also appear in the test set. Rerun under our protocol, the teacher closes only 5 of the 39 host test cases labelled closed, so host labels are not reproducible and host results appear only in Table~\ref{tab:app-host}.

\paragraph{Trajectories and audit.}
Stage 1 asks the teacher (GPT-6 Sol) for a plan of 3--12 rounds (4--16 for hosts), each with a note, 1--12 new evidence IDs and a reason, and a final answer. A program assembles the dialogue, copying tool returns verbatim, and rejects plans that cite items before they are fetched, use undeclared hypotheses, cite unread items in the final answer, contradict the determined/undetermined reference, or contain withheld identifiers. Stage 2 keeps the prose and fetches, inserts exact-anchor citations, and adds a ledger with states open, favoured and weakened. The audit returns a score from $-5$ to $+5$, an action (keep, revise, skip), a hard-failure flag and five checks (source alignment, evidence grounding, active fetching, hypothesis trajectory, honest closure).
Funnel over 1,345 non-medical cases: 996 pass Stage 1, 818 pass Stage 2, 458 are kept directly and 60 after one revision of the final answer (518).
Adding 21 pilot cases gives 539; we then added 113 supplements (50 host not-closed, 7 oversized board cases, 56 cases wrongly blocked by the identifier check) and 105 failed report cases whose plans were rewritten by a second LLM and then passed the same program checks and audit, and removed 26 board cases whose brief and evidence came from different incidents, for 731.

\paragraph{Splits.}
The split (seed 20260925) is balanced by source subtype and label and keeps each report incident in one split. Host cases are split by case: 21 of 264 host machines have cases in more than one split (48 cases).

\paragraph{OOD test set.}
83 reports (NTSB highway 39, NTSB pipeline 25, TSB Canada pipeline 19, mostly determined), 20 medical cases relabelled by whether the record contains the confirming finding (8 determined, 12 not), 18 microservice faults (one per system and fault type), and 20 Tennessee Eastman runs relabelled by whether the data can identify the fault; 109 should close and 32 should not. Domain predicts label: answering each domain's majority label, read from the test answers, gives 90.3 balanced accuracy, so we report domains separately (Table~\ref{tab:app-ood-domain}).

\paragraph{Counterfactual versions.}
The grounds of a case are the evidence items the teacher's final answer cites as support for its conclusion. Versions remove 40\%, 75\% or 100\% of the grounds; the control removes as many items never cited by the teacher. $E_1$ is never removed, identifiers are not renumbered, and the brief is regenerated with the removed items dropped from the index. A leak check flags removed items whose content is likely restated elsewhere; no edits are made, and the 20 cases without a flagged item in either the 100\% or the control version form the leakage-clean subset.

\paragraph{Reference audit.}
Of 247 references checked, about 10 (4\%) contain an extraction error, such as a conclusion field holding a safety notice. We kept them unchanged, since every stage of the pipeline used the same references.

\section{Training details}
\label{app:training}

\paragraph{SFT.}
Qwen3.5-9B, LoRA rank 16, $\alpha=32$, dropout 0, all linear layers; AdamW, peak learning rate $5\times10^{-5}$ in epoch 1 and $2\times10^{-5}$ in epoch 2 (cosine, 8 warm-up steps), gradient clipping 1.0; batch size 1 per GPU on two H100 GPUs, no packing; 32,768-token sequences in bf16; 273 steps per epoch. The training split has 6.0M tokens, 1.4M of them supervised. Validation cross-entropy after epochs 1--4 is 0.554, 0.537, 0.532 and 0.532; we chose epoch 2 before any test-set use for the reasons in \S\ref{sec:training}. We also computed token loss on a 33-case pilot set of teacher trajectories from the OOD domains (lowest at epoch 3, 0.548 vs.\ 0.549 at epoch 2). All 33 cases belong to the OOD test set, so we do not count this as a selection criterion and report the overlap here. Inference uses vLLM with the unmerged adapter; we verified that the adapter is active by comparing token losses with the training framework.

\paragraph{RLVR.}
Synchronous GRPO: both GPUs generate with vLLM, which then sleeps while two-GPU data-parallel training runs, and the new adapter is hot-swapped with a loss check. Each step draws 22 cases (8 rollouts each, at least 18 usable groups) and makes 4 AdamW updates; advantages are standardised within groups and groups with identical rewards are dropped; the loss is averaged over sampled assistant tokens. Rollouts use temperature 1.0 in non-thinking mode, 15 turns, 4,096 tokens per turn and 12,000 per case. The training pool has 544 cases (one training case lacked a complete evidence store). Format penalties: no citation $-0.25$; citing unread or non-existent items $-0.5$; each turn beyond six $-0.1$ (capped at $-1$); each repeated item beyond ten $-0.1$ (capped at $-1$); each malformed call $-0.2$ (capped at $-1$); missing marker $-2$.

\paragraph{Reward variants.}
The two judged variants add $(2q_{\text{conc}}-1)$ for closed answers and $\tfrac12(2q_{\text{conc}}-1)+\tfrac12(2q_{\text{gap}}-1)$ for unclosed ones, where $q_{\text{conc}}$ and $q_{\text{gap}}$ come from the checklist of \S\ref{sec:task} answered by GPT-6 Luna during training; ``always counted'' adds this to every rollout, ``only if decision right'' only when $D=+1$. Other runs not in the main tables: a first design that also rewarded evidence coverage and judged hypothesis updates (stopped at step 17: validation accuracy unchanged, and the hypothesis judge rewarded not updating); the always-counted variant with source-weighted sampling (stopped at step 15: closure drifted along source priors); and a run on non-host cases plus evidence-removed variants (stopped at step 28; only screened).

\paragraph{Checkpoint selection.}
Checkpoints at steps 10--50 were screened on 80 cases $\times$ 3 samples (30 validation, 30 test of which 22 non-host, 20 OOD) and a 20-case counterfactual set; step 30 was chosen for accuracy with a counterfactual effect on par with \sft{} on that set. Table~\ref{tab:app-excl} removes the screening cases. Because 9 of the 10 non-host test cases the source-only rule gets wrong were among them, the rule scores 98.3 on the rest; these numbers are therefore not evidence of robustness.

\begin{table}[h]
\centering
\small
\caption{Test non-host cases not used in checkpoint screening (42 cases: 12 should close, 30 should not). Balanced accuracy.}
\label{tab:app-excl}
\begin{tabular}{lc}
\toprule
Model & Balanced \\
\midrule
\sft{} & 68.6 \\
RL: checklist always counted (step 10) & 57.2 \\
RL: checklist if decision right (step 30) & 72.5 \\
\rlvr{} & 86.4 \\
GPT-6 Sol (teacher, 1 sample) & 85.0 \\
Source-only rule & 98.3 \\
\bottomrule
\end{tabular}
\end{table}

\section{Evaluation protocol and judges}
\label{app:judges}

\paragraph{Protocol.}
Local models run in vLLM (Qwen3.5-9B models with the non-thinking sampling recommended by the model card), 3 samples per case, 15 turns, 4,096 tokens per turn, 12,000 generated tokens per case and a 32k context. API models run through OpenRouter with provider-default sampling and reasoning, reasoning content replayed across turns, 64,000 output tokens per turn and no per-case cap, so that reasoning is not truncated. A run without \closed{} or \notclosed{} counts as wrong on both sides. Confidence intervals are 95\% case bootstraps (5,000 draws) stratified by label (and by source $\times$ label for within-source accuracy), shared across conditions so that differences are paired.

\paragraph{Judges.}
GPT-6 Sol judges the first sample of every condition, seeing the final answer, the titles of read and unread items, the text of cited items and the reference (hidden from the investigator), but not the condition name. Hedged statements (``X was not established'') are not counted as claims. When an answer makes no causal claim, the supported-claims item does not apply and the conclusion score averages the other four items. Because Sol is also the teacher, its own investigations are judged by Claude Sonnet 5 (Table~\ref{tab:app-content-sonnet}), and 50 stratified requests were re-judged by Sonnet 5 and Grok 4.7 (Table~\ref{tab:app-cross-judge}): conclusion scores correlate at 0.71--0.82, agreement on the cause item is 88\% for every pair, and although absolute overstatement rates differ across judges (70\%, 28\% and 72\% overall), all three judges find trained models overstate far less than untrained ones.

\section{Additional results}
\label{app:more}

Tables~\ref{tab:app-main}--\ref{tab:app-ood-domain} repeat the main results with confidence intervals. On boards, paired differences in balanced accuracy are +8.7 [$-$12.2, +31.1] for \rlvr{} minus Gemini and +8.5 [$-$1.3, +18.3] for \rlvr{} minus \sft{}. The tables also give counterfactual results for all conditions and versions (Tables~\ref{tab:app-counterfactual}, \ref{tab:app-cf-appendix}), conclusion quality with all columns (Table~\ref{tab:app-content}), host and all-112 results (Table~\ref{tab:app-host}), and OOD per domain (Table~\ref{tab:app-ood-domain}).

\section{Own fetching and long case files}
\label{app:long}

\begin{figure}[h]
\centering
\includegraphics[width=0.8\linewidth]{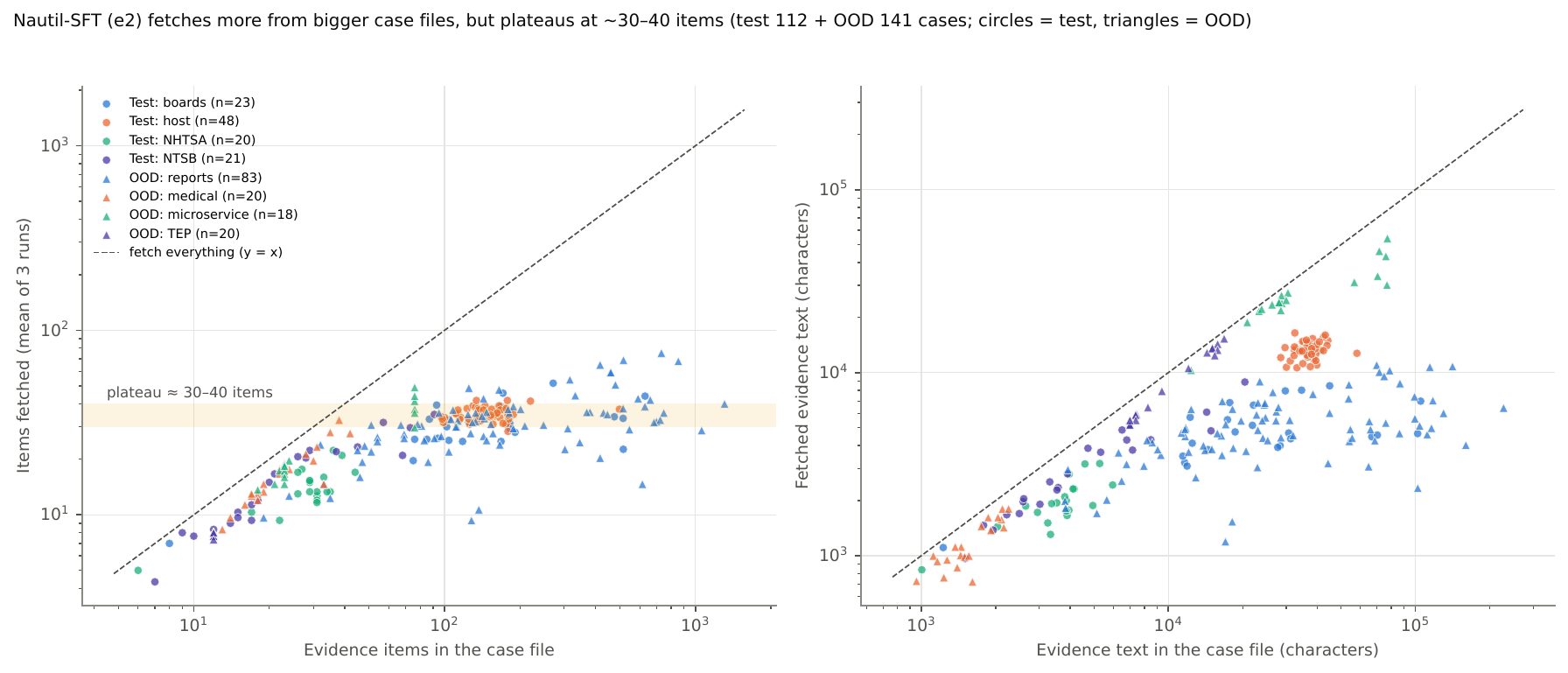}
\caption{Evidence fetched by \sft{} against case size (test and OOD, mean of 3 samples). Fetching grows with the case but levels off at 30--40 items beyond about 100 items per case.}
\label{fig:fetch}
\end{figure}

\sft{} fetches about two thirds of the items in cases with up to 50 items, but only 30--40 items (7--23\%) in cases with 100 to over 1,000 items (Figure~\ref{fig:fetch}). Table~\ref{tab:app-long-cases} compares own fetching with giving the whole file in the first message: on the 20 largest files the untrained model closes every case when given everything (an always-close pattern), while \rlvr{} closes more of the should-close cases with its own fetching (93\% vs.\ 61\%); the main cause is matched more often when everything is given, because own fetching reads only a small share of the largest files. The v2 comparison on 16 medium-long cases does not separate the two modes.


\providecommand{\oursmodel}{Nautil}
\providecommand{\nautilfit}[1]{\resizebox{\ifdim\width>\linewidth\linewidth\else\width\fi}{!}{#1}}

\begin{table*}[h]
\centering
\small
\setlength{\tabcolsep}{3.5pt}
\caption{Main results. Test set, non-host cases (23 should close / 41 should not), and OOD (109 / 32). Close$\to$C: share of should-close cases the model closed; Not$\to$N: share of should-not-close cases it left open; Bal.: their mean. Within-src: balanced accuracy computed inside each source (boards, NHTSA, NTSB) and averaged, which is exactly 50 for any rule that only looks at the source. No-mark / Fetch-err / Turns refer to the test non-host runs. A run with no closure marker counts as wrong on both sides. 9B models: mean of 3 samples per case; $^{1}$one sample per case (API models, provider-default sampling, no reasoning cap).}
\label{tab:app-main}
\nautilfit{%
\begin{tabular}{lccccccccc}
\toprule
& \multicolumn{4}{c}{Test, non-host} & \multicolumn{2}{c}{OOD} & \multicolumn{3}{c}{Runs (test non-host)} \\
\cmidrule(lr){2-5}\cmidrule(lr){6-7}\cmidrule(lr){8-10}
Model & Close$\to$C & Not$\to$N & Bal. & Within-src & Bal. & Within-dom & No-mark & Fetch-err & Turns \\
\midrule
Qwen3.5-9B (base) & 81.2 & 18.7 & 49.9 {\scriptsize[43.7, 56.1]} & 48.2 {\scriptsize[42.3, 54.4]} & 50.6 {\scriptsize[45.6, 56.3]} & 46.7 & 9.4 & 23.4 & 5.4 \\
\ + ReAct prompt & 63.8 & 31.7 & 47.7 {\scriptsize[39.0, 56.3]} & 47.9 {\scriptsize[39.5, 56.0]} & 44.7 {\scriptsize[38.2, 51.7]} & 45.9 & 19.8 & 20.3 & 8.4 \\
\ + ledger prompt & 92.8 & 17.9 & 55.3 {\scriptsize[49.8, 61.2]} & 56.9 {\scriptsize[52.2, 62.2]} & 56.9 {\scriptsize[50.0, 64.3]} & 56.8 & 3.6 & 10.4 & 5.5 \\
Abstain-R1 (3B) & 8.7 & 47.2 & 27.9 {\scriptsize[22.6, 33.7]} & 25.3 {\scriptsize[20.0, 30.8]} & 39.1 {\scriptsize[34.3, 43.8]} & 37.4 & 43.8 & 15.6 & 3.7 \\
Qwen2.5-3B-Instruct & 10.1 & 68.3 & 39.2 {\scriptsize[33.4, 45.4]} & 40.9 {\scriptsize[32.9, 49.6]} & 31.9 {\scriptsize[25.7, 38.1]} & 27.7 & 32.3 & 12.5 & 2.4 \\
\midrule
Gemini 3.8 Flash$^{1}$ & 100.0 & 58.5 & 79.3 {\scriptsize[72.0, 86.6]} & 77.0 {\scriptsize[68.5, 86.1]} & 68.9 {\scriptsize[59.9, 77.9]} & 59.1 & 0.0 & 0.0 & 4.4 \\
\midrule
\oursmodel{}-SFT & 44.9 & 93.5 & 69.2 {\scriptsize[62.4, 76.1]} & 60.4 {\scriptsize[51.7, 68.8]} & 61.7 {\scriptsize[56.4, 66.6]} & 52.1 & 5.2 & 9.4 & 5.5 \\
\oursmodel{}-RLVR & 81.2 & 85.4 & 83.3 {\scriptsize[74.8, 90.5]} & 74.1 {\scriptsize[67.3, 80.4]} & 67.0 {\scriptsize[60.0, 73.7]} & 59.3 & 0.0 & 0.5 & 4.6 \\
GPT-6 Sol (teacher ref.)$^{1}$ & 78.3 & 85.4 & 81.8 {\scriptsize[71.6, 91.5]} & 73.0 {\scriptsize[56.3, 89.3]} & n/a & n/a & 0.0 & 21.9 & 5.6 \\
\midrule
Source-only rule & 78.3 & 87.8 & 83.0 {\scriptsize[73.1, 92.0]} & 50.0 {\scriptsize[50.0, 50.0]} & 90.3 {\scriptsize[83.9, 95.4]}$^\dagger$ & 50.0$^\dagger$ & -- & -- & -- \\
\bottomrule
\end{tabular}}
\par\vspace{2pt}
\parbox{\linewidth}{\footnotesize GPT-6 Sol generated the SFT training trajectories and its original verdicts define the non-host test labels; its row is a teacher reference, not a competitor, and it was not run on OOD. Source-only rule: boards $\to$ close, NHTSA/NTSB $\to$ not close (training-set majority label of each source); it reads no evidence. $^\dagger$On OOD the rule row uses each domain's majority label among the test answers: an oracle upper bound, not a usable baseline.}
\end{table*}

\begin{table*}[h]
\centering
\small
\setlength{\tabcolsep}{3.5pt}
\caption{Counterfactual grounding test (V2). Closure rate (\%) on the intact file, on a control that removes as many non-grounding items as the 100\% version removes, and after removing 100\% of the evidence the teacher's conclusion rests on. Removed$-$control: closure-rate difference (more negative = the verdict follows the grounds). CF-Bal.: balanced accuracy over the two versions, per case (closed on control + explicitly not closed on removed)/2; a run without a closure marker counts as wrong on both sides, as in Table~\ref{tab:app-main}. vs SFT: paired difference against \oursmodel{}-SFT. Leakage-clean: neither the 100\% nor the control version removes an item flagged as restated elsewhere.}
\label{tab:app-counterfactual}
\nautilfit{%
\begin{tabular}{lccccccc}
\toprule
Model & Intact & Control & $-$100\% & Removed$-$control & vs SFT & CF-Bal. & vs SFT \\
\midrule
\multicolumn{8}{l}{\textit{Non-host cases (23; headline)}} \\
Qwen3.5-9B (base) & 87.0 & 92.8 & 87.0 & $-5.8$ {\scriptsize[$-15.9$, $+4.3$]} & $+20.3$ {\scriptsize[$0.0$, $+39.1$]} & 47.8 {\scriptsize[44.9, 50.7]} & $-14.5$ {\scriptsize[$-23.9$, $-5.1$]} \\
Gemini 3.8 Flash$^{\ddagger}$ & 97.1 & 100.0 & 92.8 & $-7.2$ {\scriptsize[$-17.4$, $0.0$]} & $+18.8$ {\scriptsize[$-1.4$, $+37.7$]} & 53.6 {\scriptsize[50.0, 58.7]} & $-8.7$ {\scriptsize[$-18.8$, $+2.9$]} \\
\oursmodel{}-SFT & 50.7 & 52.2 & 26.1 & $-26.1$ {\scriptsize[$-43.5$, $-8.7$]} & -- & 62.3 {\scriptsize[53.6, 71.0]} & -- \\
\oursmodel{}-RLVR & 75.4 & 76.8 & 60.9 & $-15.9$ {\scriptsize[$-30.4$, $-1.4$]} & $+10.1$ {\scriptsize[$-10.1$, $+29.0$]} & 58.0 {\scriptsize[50.7, 65.2]} & $-4.3$ {\scriptsize[$-14.5$, $+5.8$]} \\
GPT-6 Sol (teacher ref.)$^{1}$ & 87.0 & 82.6 & 34.8 & $-47.8$ {\scriptsize[$-69.6$, $-26.1$]} & $-21.7$ {\scriptsize[$-44.9$, $+2.9$]} & 71.7 {\scriptsize[60.9, 82.6]} & $+9.4$ {\scriptsize[$-2.2$, $+21.0$]} \\
\midrule
\multicolumn{8}{l}{\textit{All teacher-closed test cases (62, incl. 39 host)}} \\
Qwen3.5-9B (base) & 90.9 & 94.1 & 92.5 & $-1.6$ {\scriptsize[$-6.5$, $+3.2$]} & $+15.1$ {\scriptsize[$+5.4$, $+25.3$]} & 48.1 {\scriptsize[46.2, 49.7]} & $-7.8$ {\scriptsize[$-12.9$, $-3.0$]} \\
Gemini 3.8 Flash$^{\ddagger}$ & 95.7 & 96.8 & 92.5 & $-4.3$ {\scriptsize[$-12.4$, $+3.2$]} & $+12.4$ {\scriptsize[$-0.5$, $+24.2$]} & 52.2 {\scriptsize[48.4, 56.2]} & $-3.8$ {\scriptsize[$-9.9$, $+3.0$]} \\
\oursmodel{}-SFT & 72.0 & 69.4 & 52.7 & $-16.7$ {\scriptsize[$-25.8$, $-7.5$]} & -- & 55.9 {\scriptsize[51.1, 60.8]} & -- \\
\oursmodel{}-RLVR & 88.7 & 88.2 & 79.0 & $-9.1$ {\scriptsize[$-16.1$, $-2.2$]} & $+7.5$ {\scriptsize[$-2.7$, $+17.2$]} & 54.6 {\scriptsize[51.1, 58.1]} & $-1.3$ {\scriptsize[$-6.7$, $+4.0$]} \\
GPT-6 Sol (teacher ref.)$^{1}$ & 40.3 & 40.3 & 14.5 & $-25.8$ {\scriptsize[$-37.1$, $-16.1$]} & $-9.1$ {\scriptsize[$-22.0$, $+3.2$]} & 62.1 {\scriptsize[57.3, 67.7]} & $+6.2$ {\scriptsize[$0.0$, $+12.6$]} \\
\midrule
\multicolumn{8}{l}{\textit{Leakage-clean subset (20)}} \\
Qwen3.5-9B (base) & 90.0 & 96.7 & 88.3 & $-8.3$ {\scriptsize[$-16.7$, $-1.7$]} & $+15.0$ {\scriptsize[$-5.0$, $+33.3$]} & 49.2 {\scriptsize[47.5, 50.0]} & $-12.5$ {\scriptsize[$-20.8$, $-3.3$]} \\
Gemini 3.8 Flash$^{\ddagger}$ & 88.3 & 90.0 & 96.7 & $+6.7$ {\scriptsize[$-5.0$, $+23.3$]} & $+30.0$ {\scriptsize[$+10.0$, $+46.7$]} & 46.7 {\scriptsize[38.3, 52.5]} & $-15.0$ {\scriptsize[$-23.4$, $-5.0$]} \\
\oursmodel{}-SFT & 61.7 & 70.0 & 46.7 & $-23.3$ {\scriptsize[$-40.0$, $-5.0$]} & -- & 61.7 {\scriptsize[52.5, 70.0]} & -- \\
\oursmodel{}-RLVR & 83.3 & 83.3 & 78.3 & $-5.0$ {\scriptsize[$-16.7$, $+6.7$]} & $+18.3$ {\scriptsize[$-0.0$, $+36.7$]} & 52.5 {\scriptsize[46.7, 58.3]} & $-9.2$ {\scriptsize[$-18.3$, $+0.0$]} \\
GPT-6 Sol (teacher ref.)$^{1}$ & 55.0 & 55.0 & 20.0 & $-35.0$ {\scriptsize[$-55.0$, $-15.0$]} & $-11.7$ {\scriptsize[$-33.3$, $+10.0$]} & 67.5 {\scriptsize[57.5, 77.5]} & $+5.8$ {\scriptsize[$-5.0$, $+16.7$]} \\
\midrule
\multicolumn{8}{l}{\textit{Leakage-clean, non-host (10)}} \\
Qwen3.5-9B (base) & 86.7 & 96.7 & 86.7 & $-10.0$ {\scriptsize[$-23.3$, $0.0$]} & $+16.7$ {\scriptsize[$-16.7$, $+46.7$]} & 48.3 {\scriptsize[45.0, 50.0]} & $-15.0$ {\scriptsize[$-30.0$, $+1.7$]} \\
Gemini 3.8 Flash$^{\ddagger}$ & 96.7 & 100.0 & 93.3 & $-6.7$ {\scriptsize[$-16.7$, $0.0$]} & $+20.0$ {\scriptsize[$-10.0$, $+46.7$]} & 53.3 {\scriptsize[50.0, 58.3]} & $-10.0$ {\scriptsize[$-23.4$, $+5.0$]} \\
\oursmodel{}-SFT & 46.7 & 56.7 & 30.0 & $-26.7$ {\scriptsize[$-53.3$, $+3.3$]} & -- & 63.3 {\scriptsize[48.3, 76.7]} & -- \\
\oursmodel{}-RLVR & 73.3 & 80.0 & 70.0 & $-10.0$ {\scriptsize[$-30.0$, $+10.0$]} & $+16.7$ {\scriptsize[$-13.3$, $+43.3$]} & 55.0 {\scriptsize[45.0, 65.0]} & $-8.3$ {\scriptsize[$-21.7$, $+6.7$]} \\
GPT-6 Sol (teacher ref.)$^{1}$ & 100.0 & 90.0 & 40.0 & $-50.0$ {\scriptsize[$-80.0$, $-20.0$]} & $-23.3$ {\scriptsize[$-50.0$, $+6.7$]} & 75.0 {\scriptsize[60.0, 90.0]} & $+11.7$ {\scriptsize[$-3.3$, $+25.0$]} \\
\bottomrule
\end{tabular}}
\par\vspace{2pt}
\parbox{\linewidth}{\footnotesize How to read: Intact and Control should be high, $-$100\% should be low; the more negative Removed$-$control, the more closure follows the grounds. $^{1}$One sample per case and version. $^{\ddagger}$Three samples per case and version on the 23 non-host cases and one on the 39 host cases (per-case mean, then mean over cases), so the all-62 and leakage-clean-20 blocks mix the two; the one-sample Gemini row is in Table~\ref{tab:app-cf-appendix}. GPT-6 Sol is the teacher: the removed grounds are the items its own teacher answer cited, and it closes only 40\% of intact files (13\% of host files), which limits the room to drop.}
\end{table*}

\begin{table*}[h]
\centering
\small
\setlength{\tabcolsep}{3.5pt}
\caption{Appendix. Counterfactual closure rates for all versions on the 23 non-host cases, and removed$-$control on the 23 non-host / all 62 / 20 leakage-clean cases, for all conditions. The two 3B models close only 9--14\% of intact files and often produce no marker, so their rows carry little information.}
\label{tab:app-cf-appendix}
\nautilfit{%
\begin{tabular}{lcccccccc}
\toprule
& \multicolumn{5}{c}{Closure rate, non-host 23} & \multicolumn{3}{c}{Removed$-$control} \\
\cmidrule(lr){2-6}\cmidrule(lr){7-9}
Model & Intact & $-$40\% & $-$75\% & $-$100\% & Control & Non-host 23 & All 62 & Clean 20 \\
\midrule
Qwen3.5-9B (base) & 87.0 & 88.4 & 79.7 & 87.0 & 92.8 & $-5.8$ {\scriptsize[$-15.9$, $+4.3$]} & $-1.6$ {\scriptsize[$-6.5$, $+3.2$]} & $-8.3$ {\scriptsize[$-16.7$, $-1.7$]} \\
Gemini 3.8 Flash$^{\ddagger}$ & 97.1 & 97.1 & 92.8 & 92.8 & 100.0 & $-7.2$ {\scriptsize[$-17.4$, $0.0$]} & $-4.3$ {\scriptsize[$-12.4$, $+3.2$]} & $+6.7$ {\scriptsize[$-5.0$, $+23.3$]} \\
\oursmodel{}-SFT & 50.7 & 39.1 & 29.0 & 26.1 & 52.2 & $-26.1$ {\scriptsize[$-43.5$, $-8.7$]} & $-16.7$ {\scriptsize[$-25.8$, $-7.5$]} & $-23.3$ {\scriptsize[$-40.0$, $-5.0$]} \\
\oursmodel{}-RLVR & 75.4 & 82.6 & 60.9 & 60.9 & 76.8 & $-15.9$ {\scriptsize[$-30.4$, $-1.4$]} & $-9.1$ {\scriptsize[$-16.1$, $-2.2$]} & $-5.0$ {\scriptsize[$-16.7$, $+6.7$]} \\
GPT-6 Sol (teacher ref.)$^{1}$ & 87.0 & 78.3 & 52.2 & 34.8 & 82.6 & $-47.8$ {\scriptsize[$-69.6$, $-26.1$]} & $-25.8$ {\scriptsize[$-37.1$, $-16.1$]} & $-35.0$ {\scriptsize[$-55.0$, $-15.0$]} \\
\midrule
No hypothesis ledger & 63.8 & 55.1 & 50.7 & 53.6 & 62.3 & $-8.7$ {\scriptsize[$-21.7$, $+4.3$]} & $-6.5$ {\scriptsize[$-12.9$, $0.0$]} & $-8.3$ {\scriptsize[$-21.7$, $+5.0$]} \\
No not-closed examples & 87.0 & 87.0 & 92.8 & 85.5 & 94.2 & $-8.7$ {\scriptsize[$-17.4$, $-1.4$]} & $-2.2$ {\scriptsize[$-6.5$, $+1.6$]} & $-5.0$ {\scriptsize[$-11.7$, $+1.7$]} \\
Gated checklist (s30) & 55.1 & 44.9 & 36.2 & 37.7 & 58.0 & $-20.3$ {\scriptsize[$-30.4$, $-8.7$]} & $-8.6$ {\scriptsize[$-15.6$, $-1.1$]} & $-11.7$ {\scriptsize[$-25.0$, $+1.7$]} \\
Full checklist (s10) & 17.4 & 21.7 & 11.6 & 8.7 & 24.6 & $-15.9$ {\scriptsize[$-24.6$, $-7.2$]} & $-15.6$ {\scriptsize[$-23.1$, $-8.1$]} & $-20.0$ {\scriptsize[$-35.0$, $-6.7$]} \\
Abstain-R1 (3B) & 8.7 & 13.0 & 13.0 & 11.6 & 14.5 & $-2.9$ {\scriptsize[$-17.4$, $+11.6$]} & $-4.8$ {\scriptsize[$-12.9$, $+3.8$]} & $-11.7$ {\scriptsize[$-26.7$, $+5.0$]} \\
Qwen2.5-3B-Instruct & 8.7 & 5.8 & 5.8 & 5.8 & 4.3 & $+1.4$ {\scriptsize[$-5.8$, $+8.7$]} & $0.0$ {\scriptsize[$-5.9$, $+5.9$]} & $+5.0$ {\scriptsize[$-6.7$, $+15.0$]} \\
\midrule
Gemini 3.8 Flash$^{1}$ & 100.0 & 100.0 & 91.3 & 91.3 & 100.0 & $-8.7$ {\scriptsize[$-21.7$, $0.0$]} & $-4.8$ {\scriptsize[$-12.9$, $+3.2$]} & $+5.0$ {\scriptsize[$-10.0$, $+25.0$]} \\
\bottomrule
\end{tabular}}
\par\vspace{2pt}
\parbox{\linewidth}{\footnotesize $^{1}$One sample per case and version. $^{\ddagger}$Three samples per case and version on the 23 non-host cases, one on the 39 host cases (the row used in Table~\ref{tab:app-counterfactual}); the last row is the same model with sample 0 only, for reference. Gemini's counterfactual V1 run (different data build) gave $-11.3$ on all 62; on the 49 cases whose inputs are byte-identical in V1 and V2 it gave $-14.3$ vs.\ $-4.1$: single-sample variance, not the data build.}
\end{table*}

\begin{table*}[h]
\centering
\scriptsize
\setlength{\tabcolsep}{3.5pt}
\caption{Conclusion content on the test non-host cases, judged by GPT-6 Sol (sample 0 of each condition). Concl.: mean of the checklist items (cause, chain with partial = 0.5, not overstated, alternatives addressed, share of supported claims); when an answer states no causal claim, the supported-claims item is not applicable and the score averages the other four. Overst.: \% of answers stating as established what the evidence only suggests (lower is better). Cause: initiating cause matches the reference. Cause$\wedge\neg$Overst.: both at once. Supp.: supported share of causal claims (answers with $\geq$1 claim). Alt.: addresses the main competing explanation. 0-claim: \% answers with no causal claim. \#NC: number of not-closed answers; Gap, G1 (names specific missing evidence), G2 (it is really absent from the file), G3 (matches why the reference is open) are computed on those. Answers without a closure marker were not judged.}
\label{tab:app-content}
\nautilfit{%
\begin{tabular}{lcccccccccccccc}
\toprule
Model & Scored & Concl. & Overst. & Cause & Cause$\wedge$$\neg$Overst. & Chain & Supp. & Alt. & 0-claim & \#NC & Gap & G1 & G2 & G3 \\
\midrule
Qwen3.5-9B (base) & 59/64 & 0.30 {\scriptsize[0.26, 0.34]} & 97 {\scriptsize[91, 100]} & 54 {\scriptsize[42, 67]} & 3 {\scriptsize[0, 9]} & 0.48 & 26 & 19 & 3 & 10 & 0.80 {\scriptsize[0.65, 0.93]} & 100 & 80 & 60 \\
\ + ReAct prompt & 54/64 & 0.35 {\scriptsize[0.29, 0.42]} & 94 {\scriptsize[88, 100]} & 56 {\scriptsize[42, 69]} & 4 {\scriptsize[0, 9]} & 0.56 & 29 & 26 & 4 & 12 & 0.70 {\scriptsize[0.56, 0.84]} & 92 & 64 & 58 \\
\ + ledger prompt & 62/64 & 0.28 {\scriptsize[0.23, 0.32]} & 95 {\scriptsize[89, 100]} & 55 {\scriptsize[42, 67]} & 3 {\scriptsize[0, 8]} & 0.42 & 21 & 16 & 2 & 9 & 0.66 {\scriptsize[0.49, 0.82]} & 100 & 56 & 44 \\
Abstain-R1 (3B) & 38/64 & 0.16 {\scriptsize[0.11, 0.21]} & 95 {\scriptsize[87, 100]} & 29 {\scriptsize[15, 43]} & 3 {\scriptsize[0, 8]} & 0.33 & 10 & 3 & 11 & 31 & 0.33 {\scriptsize[0.26, 0.42]} & 84 & 8 & 16 \\
Qwen2.5-3B-Instruct & 43/64 & 0.28 {\scriptsize[0.20, 0.36]} & 77 {\scriptsize[64, 89]} & 40 {\scriptsize[26, 54]} & 14 {\scriptsize[5, 25]} & 0.38 & 19 & 16 & 7 & 40 & 0.39 {\scriptsize[0.32, 0.46]} & 88 & 14 & 22 \\
Gemini 3.8 Flash$^{1}$ & 64/64 & 0.55 {\scriptsize[0.50, 0.60]} & 91 {\scriptsize[83, 97]} & 84 {\scriptsize[75, 92]} & 9 {\scriptsize[3, 17]} & 0.71 & 56 & 53 & 0 & 24 & 0.96 {\scriptsize[0.91, 1.00]} & 100 & 96 & 92 \\
\oursmodel{}-SFT & 63/64 & 0.58 {\scriptsize[0.52, 0.64]} & 35 {\scriptsize[25, 45]} & 56 {\scriptsize[44, 67]} & 43 {\scriptsize[31, 55]} & 0.59 & 78 & 38 & 27 & 47 & 0.38 {\scriptsize[0.28, 0.48]} & 32 & 80 & 43 \\
\oursmodel{}-RLVR & 64/64 & 0.53 {\scriptsize[0.47, 0.58]} & 38 {\scriptsize[27, 48]} & 62 {\scriptsize[50, 73]} & 36 {\scriptsize[25, 47]} & 0.51 & 76 & 17 & 31 & 43 & 0.39 {\scriptsize[0.30, 0.48]} & 30 & 54 & 51 \\
\bottomrule
\end{tabular}}
\par\vspace{2pt}
\parbox{\linewidth}{\footnotesize $^{1}$One sample per case. GPT-6 Sol does not judge its own investigations; see Table~\ref{tab:app-content-sonnet}.}
\end{table*}

\begin{table*}[h]
\centering
\scriptsize
\setlength{\tabcolsep}{3.5pt}
\caption{Conclusion content under a different judge (Claude Sonnet 5), test non-host cases, five conditions; bottom rows: paired differences against \oursmodel{}-SFT. GPT-6 Sol is scored only here, because it cannot judge itself. The judge's reference includes Sol's own original teacher answer, which favours Sol on Cause, Chain and G3. Sonnet 5 is more lenient than GPT-6 Sol on overstatement; compare rows within this table only.}
\label{tab:app-content-sonnet}
\nautilfit{%
\begin{tabular}{lcccccccccccccc}
\toprule
Model & Scored & Concl. & Overst. & Cause & Cause$\wedge$$\neg$Overst. & Chain & Supp. & Alt. & 0-claim & \#NC & Gap & G1 & G2 & G3 \\
\midrule
Qwen3.5-9B (base) & 59/64 & 0.38 {\scriptsize[0.32, 0.44]} & 81 {\scriptsize[71, 90]} & 47 {\scriptsize[34, 60]} & 10 {\scriptsize[3, 19]} & 0.33 & 62 & 27 & 3 & 10 & 0.70 {\scriptsize[0.51, 0.88]} & 100 & 70 & 40 \\
Gemini 3.8 Flash$^{1}$ & 64/64 & 0.75 {\scriptsize[0.69, 0.80]} & 55 {\scriptsize[42, 67]} & 89 {\scriptsize[81, 95]} & 42 {\scriptsize[31, 55]} & 0.76 & 86 & 77 & 5 & 24 & 0.97 {\scriptsize[0.94, 1.00]} & 100 & 100 & 92 \\
\oursmodel{}-SFT & 63/64 & 0.64 {\scriptsize[0.58, 0.69]} & 11 {\scriptsize[5, 19]} & 68 {\scriptsize[57, 79]} & 63 {\scriptsize[52, 75]} & 0.50 & 96 & 30 & 40 & 47 & 0.45 {\scriptsize[0.34, 0.56]} & 40 & 79 & 49 \\
\oursmodel{}-RLVR & 64/64 & 0.55 {\scriptsize[0.50, 0.60]} & 20 {\scriptsize[12, 30]} & 61 {\scriptsize[48, 72]} & 50 {\scriptsize[39, 61]} & 0.43 & 89 & 14 & 36 & 43 & 0.38 {\scriptsize[0.28, 0.50]} & 33 & 64 & 49 \\
GPT-6 Sol (teacher ref.)$^{1}$ & 64/64 & 0.87 {\scriptsize[0.83, 0.90]} & 5 {\scriptsize[0, 11]} & 91 {\scriptsize[83, 97]} & 86 {\scriptsize[77, 94]} & 0.71 & 97 & 78 & 8 & 40 & 0.91 {\scriptsize[0.85, 0.96]} & 98 & 95 & 80 \\
\midrule
Qwen3.5-9B (base) $-$ SFT & 58 & $-0.25$ {\scriptsize[$-0.33$, $-0.18$]} & $+72$ {\scriptsize[$+59$, $+84$]} & $-19$ {\scriptsize[$-36$, $-2$]} & $-53$ {\scriptsize[$-66$, $-41$]} &  & $-28$ {\scriptsize[$-38$, $-17$]} & $-2$ {\scriptsize[$-16$, $+14$]} &  &  & $+0.23$ {\scriptsize[$+0.02$, $+0.45$]} &  &  &  \\
Gemini 3.8 Flash$^{1}$ $-$ SFT & 63 & $+0.11$ {\scriptsize[$+0.04$, $+0.18$]} & $+43$ {\scriptsize[$+28$, $+56$]} & $+21$ {\scriptsize[$+8$, $+33$]} & $-21$ {\scriptsize[$-36$, $-5$]} &  & $-9$ {\scriptsize[$-17$, $-0$]} & $+46$ {\scriptsize[$+32$, $+60$]} &  &  & $+0.43$ {\scriptsize[$+0.29$, $+0.58$]} &  &  &  \\
\oursmodel{}-RLVR $-$ SFT & 63 & $-0.09$ {\scriptsize[$-0.14$, $-0.03$]} & $+10$ {\scriptsize[$0$, $+19$]} & $-6$ {\scriptsize[$-18$, $+5$]} & $-13$ {\scriptsize[$-25$, $0$]} &  & $-9$ {\scriptsize[$-19$, $+1$]} & $-16$ {\scriptsize[$-28$, $-5$]} &  &  & $-0.10$ {\scriptsize[$-0.22$, $+0.04$]} &  &  &  \\
GPT-6 Sol (teacher ref.)$^{1}$ $-$ SFT & 63 & $+0.23$ {\scriptsize[$+0.17$, $+0.28$]} & $-6$ {\scriptsize[$-15$, $+2$]} & $+22$ {\scriptsize[$+10$, $+35$]} & $+22$ {\scriptsize[$+9$, $+35$]} &  & $+2$ {\scriptsize[$-3$, $+10$]} & $+48$ {\scriptsize[$+33$, $+61$]} &  &  & $+0.48$ {\scriptsize[$+0.36$, $+0.60$]} &  &  &  \\
\bottomrule
\end{tabular}}
\end{table*}

\begin{table*}[h]
\centering
\small
\setlength{\tabcolsep}{3.5pt}
\caption{Agreement between judges on the same 50 requests (stratified sample; no Gemini answers). Item agreement in \% with Cohen's $\kappa$; conclusion-score correlation (Pearson / Spearman) and mean difference (first minus second judge). Bottom: overstatement rate by group. Untrained: not trained on our data (base, two prompts, Abstain-R1, Qwen2.5-3B); trained: all SFT/RLVR variants. The absolute overstatement rate depends strongly on the judge, the direction does not.}
\label{tab:app-cross-judge}
\nautilfit{%
\begin{tabular}{lcccccc}
\toprule
Judges & Cause & Chain & Overst. & Alt. & Concl. $r$ / $\rho$ & Mean diff. \\
\midrule
GPT-6 Sol vs Claude Sonnet 5 & 88 (0.75) & 70 (0.45) & 50 (0.15) & 78 (0.49) & 0.73 / 0.76 & $-0.14$ \\
GPT-6 Sol vs Grok 4.7 & 88 (0.74) & 74 (0.52) & 86 (0.66) & 80 (0.48) & 0.82 / 0.82 & $+0.06$ \\
Claude Sonnet 5 vs Grok 4.7 & 88 (0.74) & 88 (0.78) & 56 (0.26) & 74 (0.38) & 0.71 / 0.74 & $+0.20$ \\
\midrule
\multicolumn{7}{l}{\textit{Overstatement rate (\%) on the same 50 answers}} \\
Judge & Untrained (19) & Untrained 9B (11) & Trained (31) & All 50 & Mean Concl. & \\
GPT-6 Sol & 100 & 100 & 52 & 70 & 0.38 &  \\
Claude Sonnet 5 & 58 & 45 & 10 & 28 & 0.51 &  \\
Grok 4.7 & 95 & 91 & 58 & 72 & 0.31 &  \\
\bottomrule
\end{tabular}}
\end{table*}

\begin{table*}[h]
\centering
\small
\setlength{\tabcolsep}{3.5pt}
\caption{Appendix. Left: host cases of the test set (39 / 9), labelled by the host teacher's closure marker (no ground-truth root cause); the rule is ``always close''. These labels do not reproduce: GPT-6 Sol, re-investigating under our protocol, closed only 5 of the 39 host cases labelled should-close. Right: all 112 test cases (non-host and host labels mixed).}
\label{tab:app-host}
\nautilfit{%
\begin{tabular}{lccccccc}
\toprule
& \multicolumn{4}{c}{Host (teacher labels)} & \multicolumn{3}{c}{All 112} \\
\cmidrule(lr){2-5}\cmidrule(lr){6-8}
Model & Close$\to$C & Not$\to$N & Bal. & No-mark & Close$\to$C & Not$\to$N & Bal. \\
\midrule
Qwen3.5-9B (base) & 95.7 & 3.7 & 49.7 {\scriptsize[46.6, 53.8]} & 2.8 & 90.3 & 16.0 & 53.2 {\scriptsize[49.2, 57.6]} \\
\ + ReAct prompt & 89.7 & 3.7 & 46.7 {\scriptsize[42.7, 51.6]} & 5.6 & 80.1 & 26.7 & 53.4 {\scriptsize[47.8, 59.3]} \\
\ + ledger prompt & 90.6 & 0.0 & 45.3 {\scriptsize[41.9, 48.3]} & 4.9 & 91.4 & 14.7 & 53.0 {\scriptsize[48.4, 58.0]} \\
Abstain-R1 (3B) & 22.2 & 33.3 & 27.8 {\scriptsize[19.2, 36.5]} & 38.2 & 17.2 & 44.7 & 30.9 {\scriptsize[26.1, 35.9]} \\
Qwen2.5-3B-Instruct & 19.7 & 44.4 & 32.1 {\scriptsize[21.5, 43.0]} & 27.8 & 16.1 & 64.0 & 40.1 {\scriptsize[35.2, 44.9]} \\
\midrule
Gemini 3.8 Flash$^{1}$ & 94.9 & 0.0 & 47.4 {\scriptsize[43.6, 50.0]} & 0.0 & 96.8 & 48.0 & 72.4 {\scriptsize[65.4, 79.6]} \\
\midrule
\oursmodel{}-SFT & 79.5 & 18.5 & 49.0 {\scriptsize[40.3, 58.1]} & 1.4 & 66.7 & 80.0 & 73.3 {\scriptsize[66.9, 79.6]} \\
\oursmodel{}-RLVR & 95.7 & 7.4 & 51.6 {\scriptsize[46.2, 59.8]} & 0.0 & 90.3 & 71.3 & 80.8 {\scriptsize[74.8, 86.8]} \\
GPT-6 Sol (teacher ref.)$^{1}$ & 12.8 & 55.6 & 34.2 {\scriptsize[17.5, 50.9]} & 0.0 & 37.1 & 80.0 & 58.5 {\scriptsize[50.5, 66.8]} \\
\midrule
Source-only rule & 100.0 & 0.0 & 50.0 {\scriptsize[50.0, 50.0]} & -- & 91.9 & 72.0 & 82.0 {\scriptsize[74.7, 88.6]} \\
\bottomrule
\end{tabular}}
\end{table*}

\begin{table*}[h]
\centering
\small
\setlength{\tabcolsep}{3.5pt}
\caption{Appendix. OOD by domain: share of correct verdicts per cell (should-close cells: closed; should-not cells: left open). Case counts in parentheses. Domain and label are tightly coupled; only the medical domain has enough cases on both sides.}
\label{tab:app-ood-domain}
\nautilfit{%
\begin{tabular}{lccccccc}
\toprule
& \multicolumn{2}{c}{Reports} & \multicolumn{2}{c}{Medical} & \multicolumn{2}{c}{TEP} & Microservice \\
\cmidrule(lr){2-3}\cmidrule(lr){4-5}\cmidrule(lr){6-7}\cmidrule(lr){8-8}
Model & C (80) & N (3) & C (8) & N (12) & C (3) & N (17) & C (18) \\
\midrule
Qwen3.5-9B (base) & 83.8 & 0.0 & 87.5 & 5.6 & 77.8 & 25.5 & 94.4 \\
\ + ReAct prompt & 63.7 & 0.0 & 91.7 & 2.8 & 77.8 & 39.2 & 72.2 \\
\ + ledger prompt & 85.4 & 11.1 & 95.8 & 5.6 & 100.0 & 43.1 & 92.6 \\
Abstain-R1 (3B) & 11.7 & 66.7 & 12.5 & 61.1 & 0.0 & 72.5 & 5.6 \\
Qwen2.5-3B-Instruct & 5.8 & 33.3 & 8.3 & 50.0 & 0.0 & 68.6 & 3.7 \\
\midrule
Gemini 3.8 Flash$^{1}$ & 90.0 & 0.0 & 100.0 & 33.3 & 66.7 & 64.7 & 94.4 \\
\midrule
\oursmodel{}-SFT & 40.4 & 44.4 & 25.0 & 91.7 & 11.1 & 100.0 & 0.0 \\
\oursmodel{}-RLVR & 66.2 & 33.3 & 33.3 & 75.0 & 55.6 & 92.2 & 7.4 \\
\midrule
Source-only rule$^\dagger$ & 100.0 & 0.0 & 0.0 & 100.0 & 0.0 & 100.0 & 100.0 \\
\bottomrule
\end{tabular}}
\par\vspace{2pt}
\parbox{\linewidth}{\footnotesize $^\dagger$Oracle domain-majority rule (peeks at the label distribution). GPT-6 Sol was not run on OOD.}
\end{table*}

\begin{table*}[h]
\centering
\scriptsize
\setlength{\tabcolsep}{3.5pt}
\caption{Appendix. Long cases: the model fetching evidence itself vs.\ receiving all evidence in the first message (no tools, one turn). Judge score: GPT-6 Sol, 1--5 long-case rubric (no final answer = 1). Main cause: match / partial / miss / contradicts among judged answers. $\Delta$: paired all-at-once minus own-fetching judge score. The two modes also differ in prompt wording, turns, per-turn and context limits.}
\label{tab:app-long-cases}
\nautilfit{%
\begin{tabular}{llccccccc}
\toprule
Model & Mode & Close$\to$C & Not$\to$N & Bal. & No-mark & Score & Main cause & $\Delta$ score \\
\midrule
\multicolumn{9}{l}{\textit{Medium-long v2 (16; 8 / 8)}} \\
Qwen3.5-9B & own fetching & 79.2 & 0.0 & 39.6 {\scriptsize[31.2, 47.9]} & 7 & 2.12 & 8/28/2/5 &  \\
Qwen3.5-9B & all at once & 100.0 & 0.0 & 50.0 {\scriptsize[50.0, 50.0]} & 0 & 2.42 & 17/25/0/6 & $+0.29$ {\scriptsize[$-0.06$, $+0.67$]} \\
\oursmodel{}-SFT & own fetching & 58.3 & 50.0 & 54.2 {\scriptsize[35.4, 72.9]} & 3 & 2.58 & 13/20/11/1 &  \\
\oursmodel{}-SFT & all at once & 62.5 & 58.3 & 60.4 {\scriptsize[47.9, 72.9]} & 1 & 2.75 & 14/32/1/0 & $+0.17$ {\scriptsize[$-0.25$, $+0.58$]} \\
Abstain-R1 & own fetching & 0.0 & 41.7 & 20.8 {\scriptsize[12.5, 31.2]} & 30 & 1.54 & 0/5/27/10 &  \\
Abstain-R1 & all at once & 33.3 & 33.3 & 33.3 {\scriptsize[20.8, 45.8]} & 8 & 2.02 & 1/32/5/2 & $+0.48$ {\scriptsize[$+0.25$, $+0.71$]} \\
Qwen2.5-3B & own fetching & 4.2 & 37.5 & 20.8 {\scriptsize[10.4, 31.2]} & 23 & 1.75 & 0/12/27/6 &  \\
Qwen2.5-3B & all at once & 50.0 & 29.2 & 39.6 {\scriptsize[27.1, 52.1]} & 5 & 2.06 & 4/30/6/3 & $+0.31$ {\scriptsize[$+0.10$, $+0.54$]} \\
\midrule
\multicolumn{9}{l}{\textit{Longest 20 (v3; 18 / 2)}} \\
Qwen3.5-9B & own fetching & 70.4 & 0.0 & 35.2 {\scriptsize[27.8, 42.6]} & 16 & 1.72 & 10/19/2/14 &  \\
Qwen3.5-9B & all at once & 100.0 & 0.0 & 50.0 {\scriptsize[50.0, 50.0]} & 0 & 2.62 & 36/20/0/4 & $+0.90$ {\scriptsize[$+0.67$, $+1.12$]} \\
\oursmodel{}-RLVR & own fetching & 92.6 & 33.3 & 63.0 {\scriptsize[44.4, 81.5]} & 0 & 2.47 & 8/41/6/5 &  \\
\oursmodel{}-RLVR & all at once & 61.1 & 33.3 & 47.2 {\scriptsize[39.8, 54.6]} & 6 & 2.65 & 16/38/2/0 & $+0.18$ {\scriptsize[$0.00$, $+0.38$]} \\
\bottomrule
\end{tabular}}
\end{table*}

\section{Case studies}
\label{app:cases}

\begin{figure*}[h]
\centering
\includegraphics[width=\textwidth]{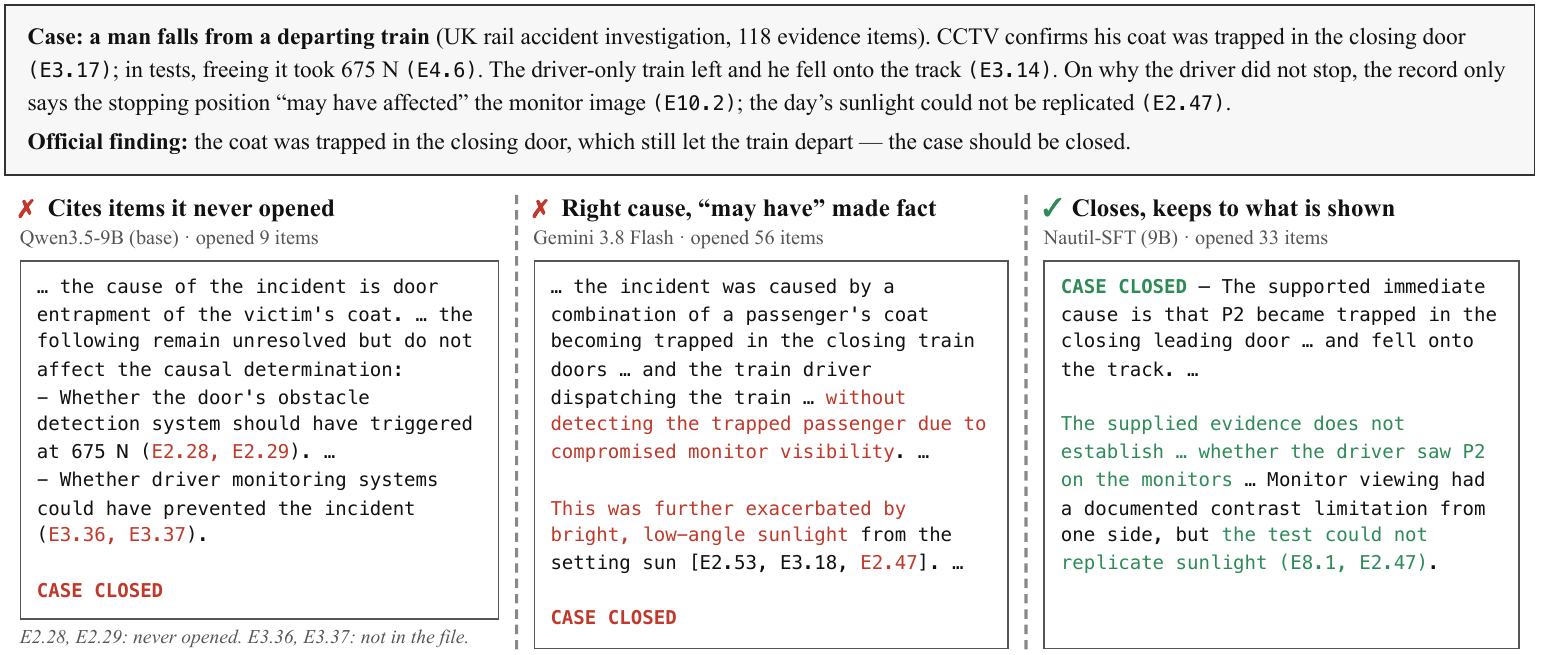}
\caption{A case that should be closed (a passenger's coat was trapped in a closing train door and the train departed); all three investigators name the trapped coat. The base model (left) cites six items it never opened, two of which do not exist; the frontier model (middle) turns hedged records of monitor contrast and sunlight into the established reason the driver missed the passenger. \sft{} (right) closes on what the CCTV and tests show and leaves what the driver saw open. Excerpts shortened (\ldots); one sample per model.}
\label{fig:caseB}
\end{figure*}

\begin{figure*}[h]
\centering
\includegraphics[width=\textwidth]{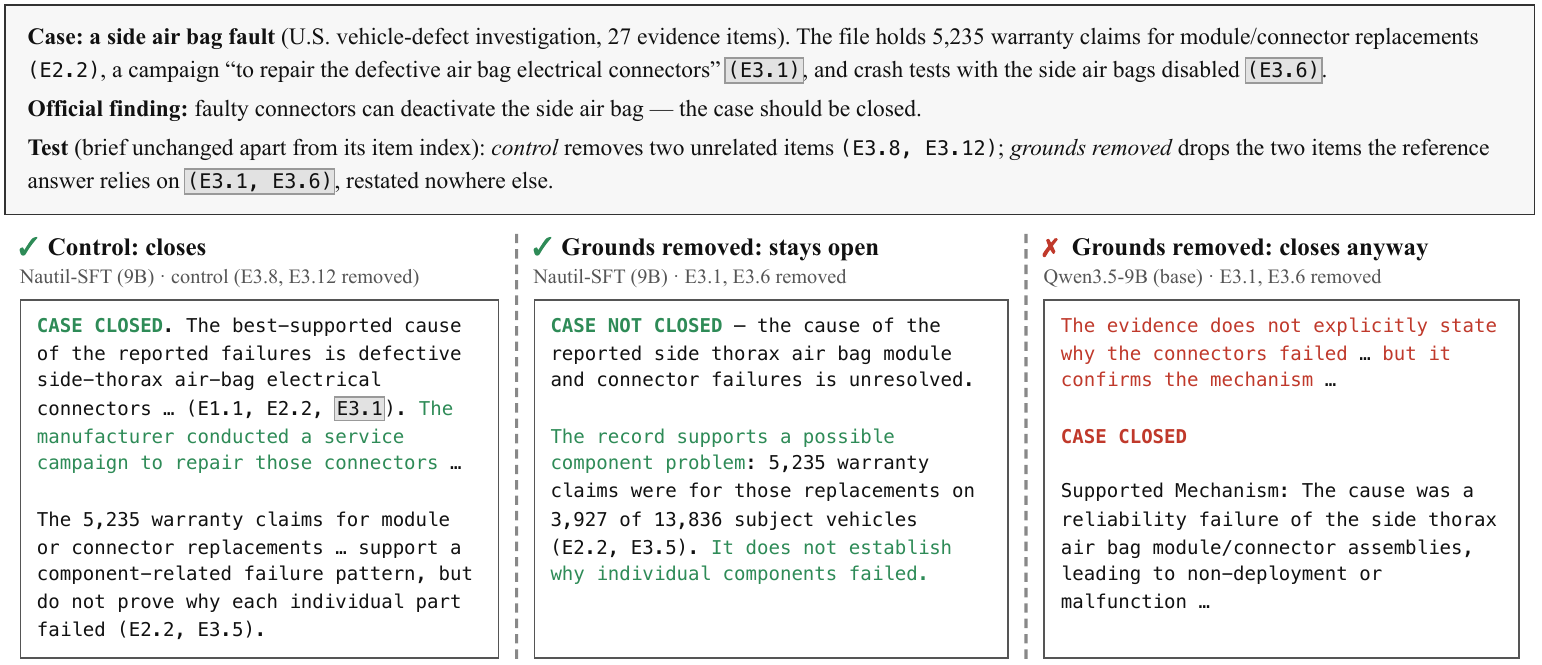}
\caption{The same case with different evidence removed; the brief is unchanged except that removed items drop out of its index. With two unrelated items removed (left), \sft{} closes on the manufacturer's campaign to repair defective connectors (E3.1); with E3.1 and the crash tests (E3.6) removed (middle), it treats the remaining warranty counts as only a possible component problem and leaves the case open. The base model (right) closes the reduced file on the warranty counts alone. Excerpts shortened (\ldots); one sample per model and version.}
\label{fig:caseC}
\end{figure*}

\begin{figure*}[h]
\centering
\includegraphics[width=\textwidth]{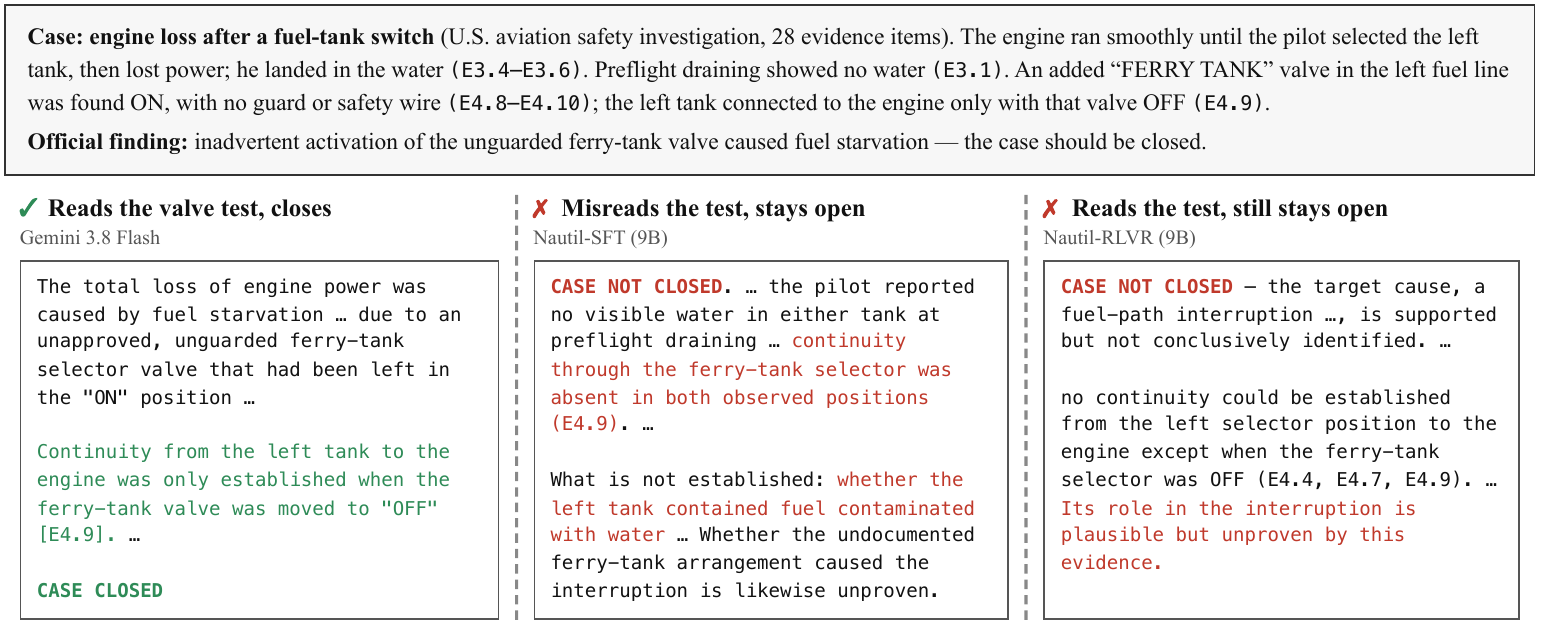}
\caption{A failure of our models on a case that should be closed: the file shows that an unguarded ferry-tank valve set to ON blocked fuel from the left tank, and the frontier model (left) reads this and closes. \sft{} (middle) misreads the continuity test (E4.9) and keeps water contamination open, although its own answer notes that preflight draining showed no water. \rlvr{} (right) reads the test correctly but still calls the valve's role unproven. Excerpts shortened (\ldots); one sample per model.}
\label{fig:caseD}
\end{figure*}

\section{Release}
\label{app:release}
The code is available at \url{https://github.com/etigerstudio/Nautil}. The \sft{} and \rlvr{} adapters are released at \url{https://huggingface.co/etigerstudio/Nautil-SFT} and \url{https://huggingface.co/etigerstudio/Nautil-RLVR}. The \ours{} report cases with their counterfactual versions and OOD set (excluding the medical cases, whose licences are mixed), the teacher trajectories and the hidden references are released at \url{https://huggingface.co/datasets/etigerstudio/Nautil}; cases from TSB Canada are under CC BY-NC 4.0 as their source requires. We will release the evaluation trajectories. Licences follow the sources listed in Appendix~\ref{app:data}. The anonymised textual host cases, their counterfactual versions and the host case extraction code will be released after internal review.

\end{document}